\def\year{2019}\relax
\documentclass[letterpaper]{article} 
\usepackage{ourstyle}  
\usepackage{times}  
\usepackage{helvet}  
\usepackage{courier}  
\usepackage{url}  
\usepackage{graphicx}  
\usepackage{hyperref}
\usepackage{url}
\usepackage{amsthm}
\usepackage{amsfonts}
\usepackage{color}
\usepackage{lipsum}  
\usepackage{multicol}

\usepackage{amsmath}
\usepackage{mathrsfs}
\usepackage{graphicx}
\usepackage{subfig}
\usepackage{floatrow}
\usepackage{algorithm}
\usepackage{algorithmic}
\usepackage{float}
\usepackage{titlesec}

\newcommand{\Cb}{\boldsymbol{C}}

\newcommand{\Ib}{\boldsymbol{I}}

\newcommand{\ub}{\boldsymbol{u}}
\newcommand{\vb}{\boldsymbol{v}}
\newcommand{\wb}{\boldsymbol{w}}
\newcommand{\xb}{\boldsymbol{x}}

\newcommand{\deltb}{\boldsymbol{\delta}}

\DeclareMathOperator*{\argmax}{arg\,max}
\makeatletter
\newcommand{\printfnsymbol}[1]{%
  \textsuperscript{\@fnsymbol{#1}}%
}
\makeatother

\begin{document}
%
\title{Interpretation of Neural Networks is Fragile}
\author{Amirata Ghorbani\thanks{Equal Contribution}\\
Dept. of Electrical Engineering\\
Stanford University\\
amiratag@stanford.edu\\
\And
Abubakar Abid\printfnsymbol{1}\\
Dept. of Electrical Engineering\\
Stanford University\\
a12d@stanford.edu\\
\And
James Zou\thanks{Corresponding author: jamesz@stanford.edu}\\
Dept. of Biomedical Data Science\\
Stanford University\\
jamesz@stanford.edu\\
}
\maketitle
\begin{abstract}
In order for machine learning to be trusted in many applications, it is critical to be able to reliably explain why the machine learning algorithm makes certain predictions. For this reason, a variety of methods have been developed recently to interpret neural network predictions by providing, for example, feature importance maps. For both scientific robustness and security reasons, it is important to know to what extent can the interpretations be altered by small systematic perturbations to the input data, which might be generated by adversaries or by measurement biases. In this paper, we demonstrate how to generate adversarial perturbations that produce perceptively indistinguishable inputs that are assigned the \emph{same} predicted label, yet have very \emph{different} interpretations. We systematically characterize the robustness of interpretations generated by several widely-used feature importance interpretation methods (feature importance maps, integrated gradients, and DeepLIFT) on ImageNet and CIFAR-10. In all cases, our experiments show that systematic perturbations can lead to dramatically different interpretations without changing the label. We extend these results to show that interpretations based on exemplars (e.g. influence functions) are similarly susceptible to adversarial attack. Our analysis of the geometry of the Hessian matrix gives insight on why robustness is a general challenge to current interpretation approaches.
\end{abstract} 

\section{Introduction}

Predictions made by machine learning algorithms play an important role in our everyday lives and can affect decisions in technology, medicine, and even the legal system \cite{rich2015machine,obermeyer2016predicting}. As algorithms become increasingly complex, explanations for why an algorithm makes certain decisions are ever more crucial. For example, if an AI system predicts a given pathology image to be malignant, then a doctor may need to know what features in the image led the algorithm to this classification. Similarly, if an algorithm predicts an individual to be a credit risk, then the lender (and the borrower) might want to know why. Therefore having interpretations for why certain predictions are made is critical for establishing trust and transparency between users and the algorithm \cite{lipton2016mythos}.

\begin{figure*}[]
\centering
\includegraphics[width=1\linewidth]{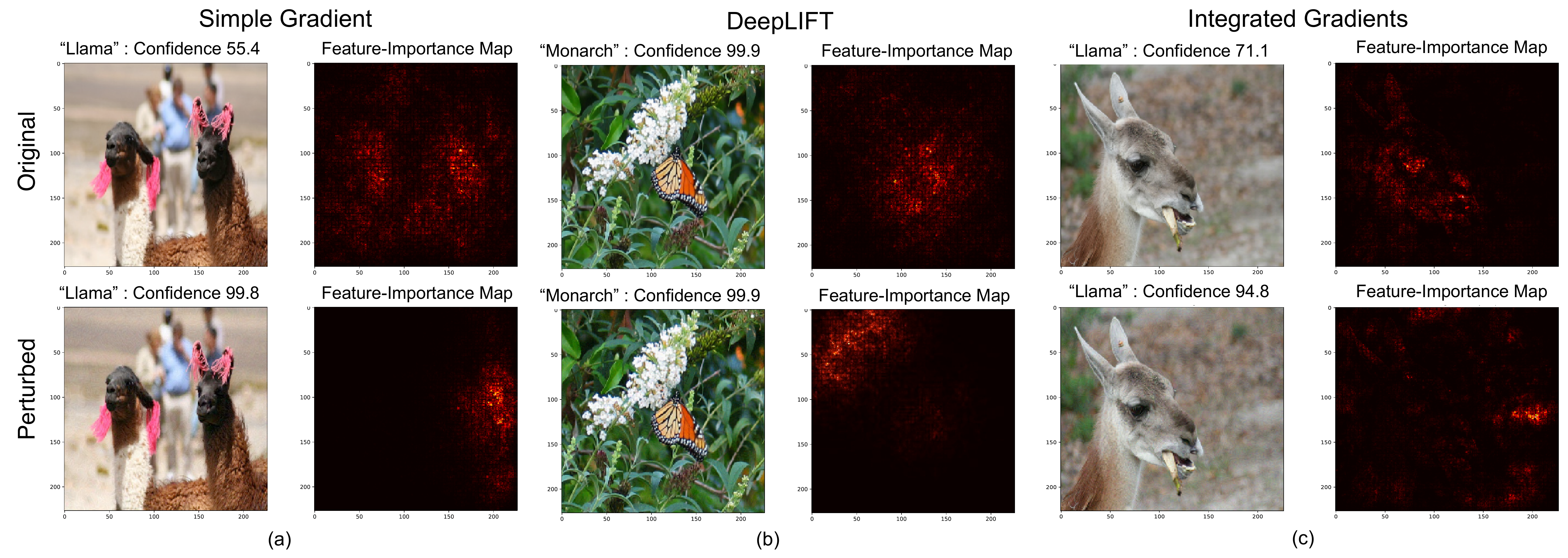} 
\caption{\textbf{ Adversarial attack against feature-importance maps.} We generate feature-importance scores, also called saliency maps, using three popular interpretation methods: (a) simple gradients, (b) DeepLIFT, and (c) integrated gradients. The \textbf{top row} shows the the original images and their saliency maps and the \textbf{bottom row} shows the perturbed images (using the center attack with $\epsilon=8$, as described in Section~\ref{sec:attack}) and corresponding saliency maps. In all three images, the predicted label does not change from the perturbation; however, the saliency maps of the perturbed images shifts dramatically to features that would not be considered salient by human perception.  
\label{fig:examples}}
\end{figure*}

Having an interpretation is not enough, however. The explanation itself must be robust in order to establish human trust. Take the pathology predictor; an interpretation method might suggest that a particular section in an image is important for the malignant classification (e.g. that section could have high scores in saliency map). The clinician might then focus on that section for investigation or treatment or even look for similar features in other patients. It would be highly disconcerting if in an extremely similar image, visually indistinguishable from the original and also classified as malignant, a very different section is interpreted as being salient for the prediction. Thus, even if the predictor is robust (both images are correctly labeled as malignant), that the interpretation is fragile would still be problematic in deployment. Furthermore, if the interpretation is used to guide interventions (e.g. location of a biopsy) by the doctor, then an interpretation that is not robust against adversarial perturbations may prove to be a security concern.       

\vspace{-0.15cm}
\paragraph{Our contributions.} 
 It is well known that the predicted \textit{labels} of deep neural networks are susceptible to adversarial attacks ~\cite{goodfellow2014explaining,kurakin2016adversarial,papernot2016limitations,moosavi2016deepfool}. In this paper, we introduce the notion of adversarial perturbations to neural network interpretation. More precisely, we define the interpretation of neural network to be \textit{fragile} if, for a given image, it is possible to generate a perceptively indistinguishable image that has 
the same prediction label by the neural network, yet is given a substantially different interpretation. We systematically investigate two classes of interpretation methods: methods that assign  importance scores to each feature (this includes simple gradients \cite{simonyan2013deep}, DeepLift \cite{Deeplift17}, and integrated gradients \cite{integrated_gradients}), as well as a method that assigns importances to each training example: influence functions \cite{koh2017understanding}. For these interpretation methods, we show how to design targeted perturbations that can lead to dramatically different interpretations across test images (Fig.~\ref{fig:examples}). 
Our findings highlight the fragility of interpretations of neural networks, which has not been carefully considered in the literature. Fragility limits how much we can trust and learn from the interpretations. It also raises a significant new security concern. Especially in medical or economic applications, users often take the interpretation of a prediction as containing causal insight (\emph{``this image is a malignant tumor because of the section with a high saliency score''}).   
An adversary could minutely manipulate the input to draw attention away from relevant features or onto his/her desired features. Such attacks might be especially hard to detect as the actual labels have not changed.  

\begin{figure}[]
\centering
\includegraphics[width=\linewidth]{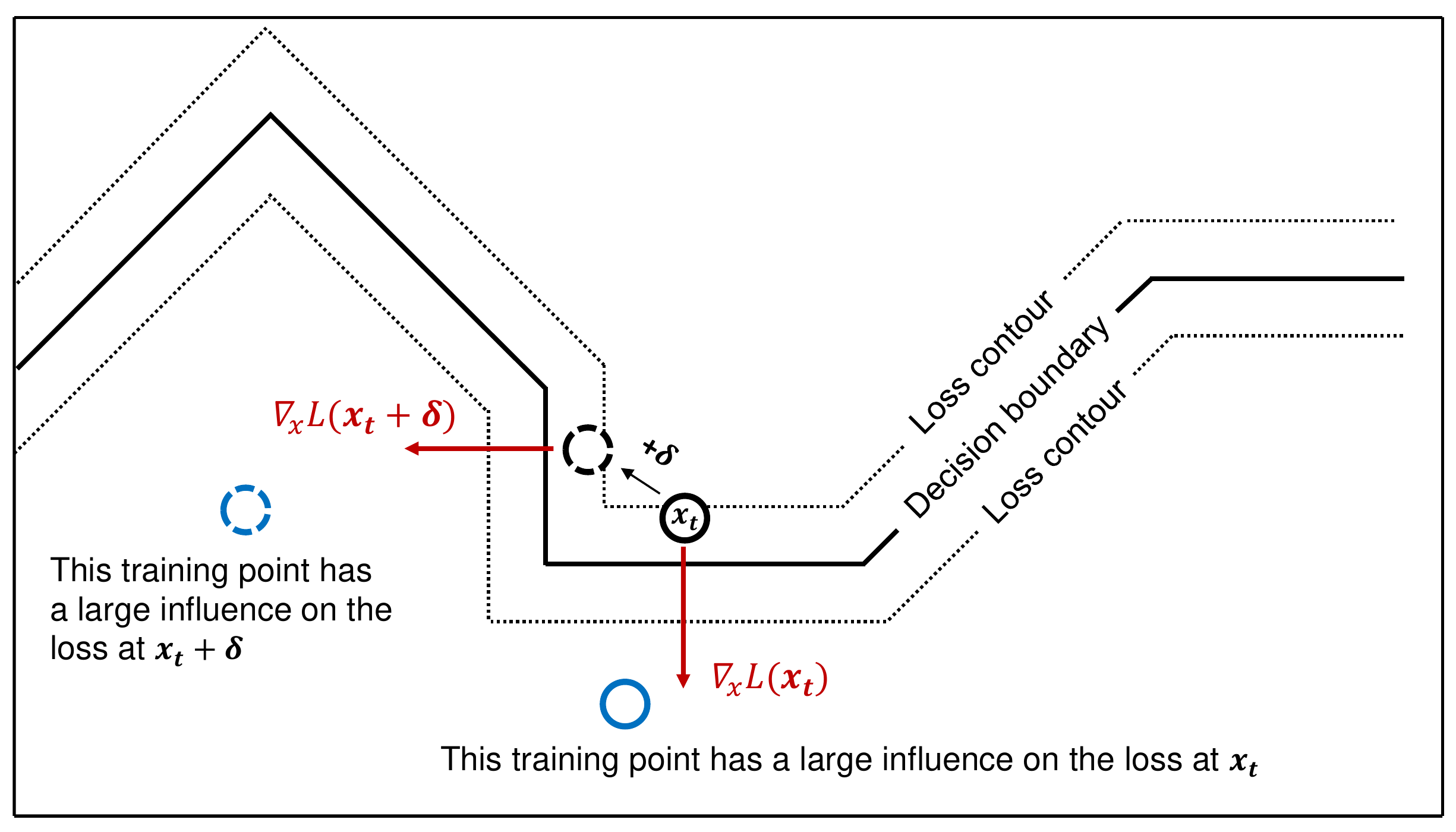}
\caption{\textbf{Intuition for why interpretation is fragile.} Consider a test example $\boldsymbol{x_t} \in \mathbb{R}^2$ (solid black circle) that is slightly perturbed to a new position $\boldsymbol{x_t}+\boldsymbol{\delta}$ in input space (dashed black dot). The contours and decision boundary corresponding to a loss function ($L$) for a two-class classification task are also shown, allowing one to see the direction of the gradient of the loss with respect to the input space. Neural networks with many parameters have decision boundaries that are roughly piecewise linear with many transitions \protect\cite{goodfellow2014explaining}. We illustrate that points near the transitions are especially fragile to interpretability-based analysis. A small perturbation to the input changes the direction of $\nabla_x L$ from being in the horizontal direction to being in the vertical direction, directly affecting feature-importance analyses. Similarly, a small perturbation to the test image changes which data point (before perturbation: solid blue, after perturbation: dashed blue), when up-weighted, has the largest influence on $L$, directly affecting exemplar-based analysis.}
\label{fig:concept}
\end{figure}

While we focus on image data here because most interpretation methods have been motivated by images, the fragility of neural network interpretation could be a much broader problem. Fig.~\ref{fig:concept} illustrates the intuition that when the decision boundary in the input feature space is complex, as is the case with deep networks, a small perturbation in the input can push the example into a region with very different loss contours. Because the feature importance is closely related to the gradient which is perpendicular to the loss contours, the importance scores can also be dramatically different. We provide additional analysis of this in Section \ref{section:geometric}.

\section{Related Works and Preliminaries}

\subsection{Related works}

 \cite{szegedy2013intriguing} first demonstrated the possibility of fooling neural networks into making different predictions for test images that are visually indistinguishable. \cite{goodfellow2014explaining} introduced the one-step Fast Gradient Sign Method (FGSM) prediction attack which was followed by more effective iterative attacks ~\cite{kurakin2016adversarial}. To quantify the perturbation size, metrics such as
 $\ell_2$  ~\cite{moosavi2016deepfool,szegedy2013intriguing}, $\ell_0$ (number of perturbed pixels) ~\cite{papernot2016limitations}, and $\ell_\infty$ ~\cite{goodfellow2014explaining} have been introduced. As it tightly controls how much individual input features can change, we followed the popular practice and adopted $\ell_\infty$. There has also been a line of work showing that networks that are robust against adversarial attacks to its predictions also have improved interpretability~\cite{ross2017improving,dong2017towards}. The focus of all of these works is on adversarial attacks against the \emph{prediction}; in contrast our work focuses on attacks on the \emph{interpretation} without changing the prediction.  

\subsection{Interpretation methods for neural networks}
Interpretation of neural network predictions is an active research area.  Post-hoc interpretability~\cite{lipton2016mythos} is one family of methods that seek to ``explain" the prediction without considering the details of black-box model's hidden mechanisms. These include methods to explain predictions in terms of the features of the test example, as well as in terms of the contribution of training examples to the test time prediction. These interpretations have gained increasing popularity, as they confer a degree of insight to human users of what the neural network might be doing \cite{lipton2016mythos}. We describe several widely-used interpretation methods in what follows.

\paragraph{Feature importance interpretation}
 This first class of methods explains predictions in terms of the relative importance of features of an input test sample. Given the sample $\xb_t \in \mathbb{R}^d$ and the network's prediction $l$, we define the score of the predicted class $S_l(\xb_t)$ to be the value of the pre-softmax layer's $l$-th  neuron. We take $l$ to be the class with the highest score; i.e. the predicted class. feature importance methods seek to find the dimensions of input data point that strongly affect the score, and in doing so, these methods assign an absolute feature importance score to each input feature. We normalize the scores for each image by the sum of the feature importance scores across the features. This ensures that any perturbations that we design change not the absolute feature saliencies (which may  preserve the ranking of different features), but their relative values. We summarize three different ways to compute normalized feature importance score, denoted by $\Ib(\xb_t)$.

\begin{itemize}

\item \textbf{Simple gradient method}
Introduced in ~\cite{baehrens2010explain} and applied to deep neural networks in ~\cite{simonyan2013deep}, the simple gradient method applies a first order linear approximation of the model to detect the sensitivity of the score to perturbing each of the input dimensions.  Given input $\xb_t \in \mathbb{R}^d$, the score is defined as:
$
\Ib(\xb_t)_j =  \rvert \nabla_{\xb} S_l(\xb_t)_{j} \lvert/\sum_{i=1}^d \rvert\nabla_{\xb} S_l(\xb_t)_{i} \lvert.
$

\item \textbf{Integrated gradients}
A significant drawback of the simple gradient method is the saturation problem discussed by \cite{Deeplift17,integrated_gradients}. Consequently, \citeauthor{integrated_gradients} introduced the integrated gradients method where the gradients of the score with respect to $M$ scaled versions of the input are summed and then multiplied by the input. Letting $\xb^0$ be the reference point and $\Delta \xb_t = \xb_t - \xb^0$, the feature importance vector is calculated by:
$
\Ib(\xb_t) = \left\lvert\frac{\Delta \xb_t}{M} \sum_{k=1}^M \nabla_{\xb}{S_l\left(\frac{k}{M} \Delta \xb_t + \xb^0\right)} \right\rvert,
$
which is then normalized for our analysis. Here the absolute value is taken for each dimension.

\item \textbf{DeepLIFT} DeepLIFT is an improved version of layer-wise relevance propagation (LRP) method \cite{bach2015pixel}. LRP methods decompose the score $S_l(\xb_t)$ backwards through the neural network. 
DeepLIFT \cite{Deeplift17} defines a reference point in the input space and propagates relevance scores proportionally to the changes in the neuronal activations from the reference. We use DeepLIFT with the Rescale rule; see \cite{Deeplift17} for details.
\end{itemize}
\paragraph{Sample Importance Interpretation}

A complementary approach to interpreting the results of a neural network is to explain the prediction of the network in terms of its \textit{training examples}, $\{(\boldsymbol{x_i}, y_i)\}$. Specifically, to ask which training examples, if up-weighted or down-weighted during training time, would have the biggest effect on the loss of the test example $(\boldsymbol{x_t}, y_t)$. \cite{koh2017understanding} proposed a method to calculate this value, called the influence, defined by the equation:
$
I(z_i, z_t) = - \nabla_\theta L(z_t, \hat{\theta})^\top H_{\hat{\theta}}^{-1} \nabla_\theta L(z_i, \hat{\theta}),
\label{eqn:influence_definition}
$
where $z_i = (\xb_i, y_i)$ , $z_t = (\xb_t, y_T)$, and $L(z, \hat{\theta})$ is the prediction loss of (training or test) data point $z$ in network with parameters $\hat{\theta}$. $H_{\hat{\theta}} = \frac{1}{n} \sum_{i=1}^{n} \nabla_\theta^2 L(z_i, \hat{\theta})$ is the empirical Hessian of the network calculated over the training examples. We calculate the influence over the entire training set $\Ib(\cdot, z_t)$.

\paragraph{Metrics for interpretation similarity}
We consider two natural metrics for quantifying the similarity between interpretations for two different images:

\begin{itemize}
    \item \textbf{Spearman's  rank order correlation}: Because interpretation methods rank all of the features or training examples in order of importance, it is natural to use the rank correlation \cite{spearman04} to compare the similarity between interpretations.
    \item \textbf{Top-$k$ intersection}: 
    In many settings, only the most important features are of explanatory interest. In such settings, we can compute the size of intersection of the $k$ most important features before and after perturbation. 
\end{itemize}

\section{Methods: Generating Perturbations}\label{sec:attack}

\paragraph{Problem statement}

For a given neural network $\mathscr{N}$ with fixed weights and a test data point $\xb_t$, the feature importance and sample importance methods produce an interpretation $\Ib (\xb_t; \mathscr{N})$. For feature importance, $\Ib (\xb_t; \mathscr{N})$ is a vector of feature scores; for influence function $\Ib (\xb_t; \mathscr{N})$ is a vector of scores for training examples.  Our goal is to devise efficient and visually imperceptible perturbations that change the interpretability of the test input while preserving the predicted label. Formally, we define the problem as:
\begin{gather*}
\argmax_{\boldsymbol{\delta}} \mathcal{D}\left(\Ib(\xb_t;\mathscr{N}),\Ib(\xb_t+\boldsymbol{\delta};\mathscr{N})\right)\\
\text{subject to:} \,\, ||\boldsymbol{\delta}||_{\infty} \le \epsilon, \\
\text{Prediction}(\xb_t+\boldsymbol{\delta}; \mathscr{N}) = \text{Prediction}(\xb_t; \mathscr{N})
\end{gather*}
where $\mathcal{D}(\cdot)$ measures the change in interpretation (e.g. how many of the top-$k$ pixels are no longer the top-$k$ pixels of the feature importance map after the perturbation) and $\epsilon > 0$ constrains the norm of the perturbation. In this paper, we carry out three kinds of input perturbations.

\paragraph{Random sign perturbation} As a baseline, each pixel is randomly perturbed by $\pm \epsilon$. This is used as a baseline with which to compare our adversarial perturbations against both feature importance and sample importance methods.

\paragraph{Iterative attacks against feature importance methods}

In Algorithm 1, we define three adversarial attacks against feature importance methods, each of which consists of taking a series of steps in the direction that maximizes a differentiable dissimilarity function between the original and perturbed interpretation. (1) The \textbf{top-$\mathbf{k}$} attack seeks to perturb the feature importance map by decreasing the relative importance of the $k$ initially most important input features.
(2) For image data, feature importance map's center of mass often captures the user's attention. The \textbf{mass-center} attack is designed to result in the maximum spatial displacement of the center of mass. (3) If the goal is to have a semantically meaningful change in feature importance map, \textbf{targeted} attack aims to increase the concentration of feature importance scores in a pre-defined region of the input image.

\begin{algorithm}[tb]
  \caption{Iterative feature importance Attacks}
  \label{alg:iterative}
  \begin{algorithmic}
  \STATE {\bfseries Input:} test image $\xb_t$, maximum norm of perturbation $\epsilon$, normalized feature importance function~$\Ib(\cdot)$,  number of iterations $P$, step size $\alpha$
   
    
    \STATE Define a dissimilarity function $D$ to measure the change between interpretations of two images:
    
    $$D(\xb_t, \xb) = \begin{cases} 
      -\sum\limits_{i \in B} {\Ib(\xb)}_i & \text{for \textbf{top-k} attack} \\
      \sum\limits_{i \in \mathcal{A}} {\Ib(\xb)}_i & \text{for \textbf{targeted} attack} \\ 
      ||\Cb(\xb) - \Cb(\xb_t)||_2 & \text{for \textbf{mass-center} attack,} 
  \end{cases}
    $$

    \STATE where $B$ is the set of the $k$ largest dimensions of $\Ib(\xb_t)$, $\mathcal{A}$ is the target region of the input image in targeted attack,  and $\Cb(\cdot)$ is the center of feature importance mass\footnote{The center of mass is defined for a $W\times H$ image as: $\Cb(\xb) = \sum_{i \in \{1,\dots,W\}} \sum_{j \in \{1,\dots,H\}} \Ib(\xb)_{i,j} [i,j]^T$}.

  \STATE Initialize $\xb^0 = \xb_t$
  \FOR{$p\in \{1, \ldots, P\}$}
  \STATE Perturb the test image in the direction of signed gradient\footnote{In ReLU networks, this gradient is 0. To attack interpretability in such networks, we replace the ReLU activation with its smooth approximation (softplus) when calculating the gradient and generate the perturbed image using this approximation. The perturbed images that result are effective adversarial attacks against the original ReLU network, as discussed in Section \ref{section:results}.} of the dissimilarity function:
  $$\xb^p =  \xb^{p-1} + \alpha \cdot \text{sign}(\nabla_{\xb} D(\xb_t, \xb^{p-1}))$$\\
  
  \STATE If needed, clip the perturbed input to satisfy the norm constraint: $||\xb^p - \xb_t||_\infty \le \epsilon$
  \ENDFOR
  \STATE Among $\{\xb^1,\dots,\xb^{P}\}$, return the element with the largest value for the dissimilarity function and the same prediction as the original test image.
  \end{algorithmic}
\end{algorithm}

\paragraph{Gradient sign attack against influence functions}

We can obtain effective adversarial images for influence functions without resorting to iterative procedures. We linearize the equation for influence functions around the values of the current inputs and parameters. If we further constrain the $L_\infty$ norm of the perturbation to $\epsilon$, we obtain an optimal single-step perturbation:


\begin{equation}
\begin{aligned}
&\boldsymbol{\delta} = \epsilon \text{sign}(\nabla_{\boldsymbol{x}_t} I(z_i, z_t)) = \\
& - \epsilon \text{sign}(\nabla_{\boldsymbol{x}_t} \nabla_\theta L(z_t, \hat{\theta})^\top \underbrace{H_{\hat{\theta}}^{-1} \nabla_\theta L(z_i, \hat{\theta})}_{\text{independent of $\boldsymbol{x}_t$}})
\label{eqn:influence_perturbation}
\end{aligned}
\end{equation}


The attack we use consists of applying the negative of the perturbation in (\ref{eqn:influence_perturbation}) to decrease the influence of the 3 most influential training images of the original test image\footnote{In other words, we generate the perturbation given by: $- \epsilon \text{sign}(\sum_{i=1}^3 \nabla_{\boldsymbol{x}_t} \nabla_\theta L(z_t, \hat{\theta})^\top H_{\hat{\theta}}^{-1} \nabla_\theta L(z_{(i)}, \hat{\theta}))$, where $z_{(i)}$ is the $i^{\text{th}}$ most influential training image of the original test image.}. Of course, this affects the influence of all of the other training images as well.

We follow the same setup for computing the influence function as was done in \cite{koh2017understanding}. Because the influence is only calculated with respect to the parameters that change during training, we calculate the gradients only with respect to parameters in the final layer of our network (InceptionNet, see Section \ref{section:results}). This makes it feasible for us to compute (\ref{eqn:influence_perturbation}) exactly, but it gives us the perturbation of the input \textit{into the final layer}, not the first layer. So, we use standard back-propagation to calculate the corresponding gradient for the input test image.

\section{Experiments \& Results}
\label{section:results}

\paragraph{Data sets and models}
For attacks against feature importance interpretation, we used ILSVRC2012 (ImageNet classification challenge data) ~\cite{ILSVRC15} and CIFAR-10 ~\cite{cifar10}. 
For the ImageNet classification data set, we used a pre-trained SqueezeNet 
model introduced by ~\cite{squeezenet16}. For the CIFAR-10 data we trained our own convolutional network (architecture in Appendix~\ref{appendix:CIFAR10-network}.)

For both data sets, the results are examined on feature importance scores obtained by simple gradient, integrated gradients, and DeepLIFT methods. For DeepLIFT, we used the pixel-wise and the channel-wise mean images as the CIFAR-10 and ImageNet reference points respectively.  For the integrated gradients method, the same references were used with parameter $M=100$. We ran all iterative attack algorithms for $P=300$ iterations with step size $\alpha=0.5$. (To show the performance success against methods that are not directly gradient-based, we also ran a smaller experiment of 100 Imagenet examples for the Deep Taylor Decomposition method ~\cite{DTD} to show the attack method's success  Results are reflected in Appendix~\ref{appendix:results}).

To evaluate our adversarial attack against influence functions, we followed a similar experimental setup to that of the original authors: we trained an InceptionNet v3 with all but the last layer frozen (the weights were pre-trained on ImageNet and obtained from Keras
). The last layer was trained on a binary flower classification task (\textbf{roses} vs. \textbf{sunflowers}), using a data set consisting of 1,000 training images\footnote{adapted from: \url{https://goo.gl/Xgr1a1}}. This data set was chosen because it consisted of images that the network had not seen during pre-training on ImageNet. The network achieved a validation accuracy of 97.5\%.


\paragraph{Results for attacks against feature importance scores}
\label{subsec:results_exp} From the ImageNet test set, 512 correctly-classified images were randomly sampled for evaluation. Examples of the mass-center attack against feature importance scores obtained by the three mentioned methods are presented in Fig.~\ref{fig:examples}. Examples of targeted attacks,  whose goal is to change the semantic meaning of the interpretation are depicted in Fig.~\ref{fig:semantic} and also in Appendix~\ref{appendix:semantic}. Further representative examples of top-k and mass center attacks are found in Appendix~\ref{appendix:examples}. Appendix~\ref{appendix:metrics} provides examples of how the decrease in rank order correlation and top-1000 intersection relate to visual changes in the feature importance maps.



\begin{figure*}[ht]
\centering
\subfloat[]{\includegraphics[width=0.26\linewidth]{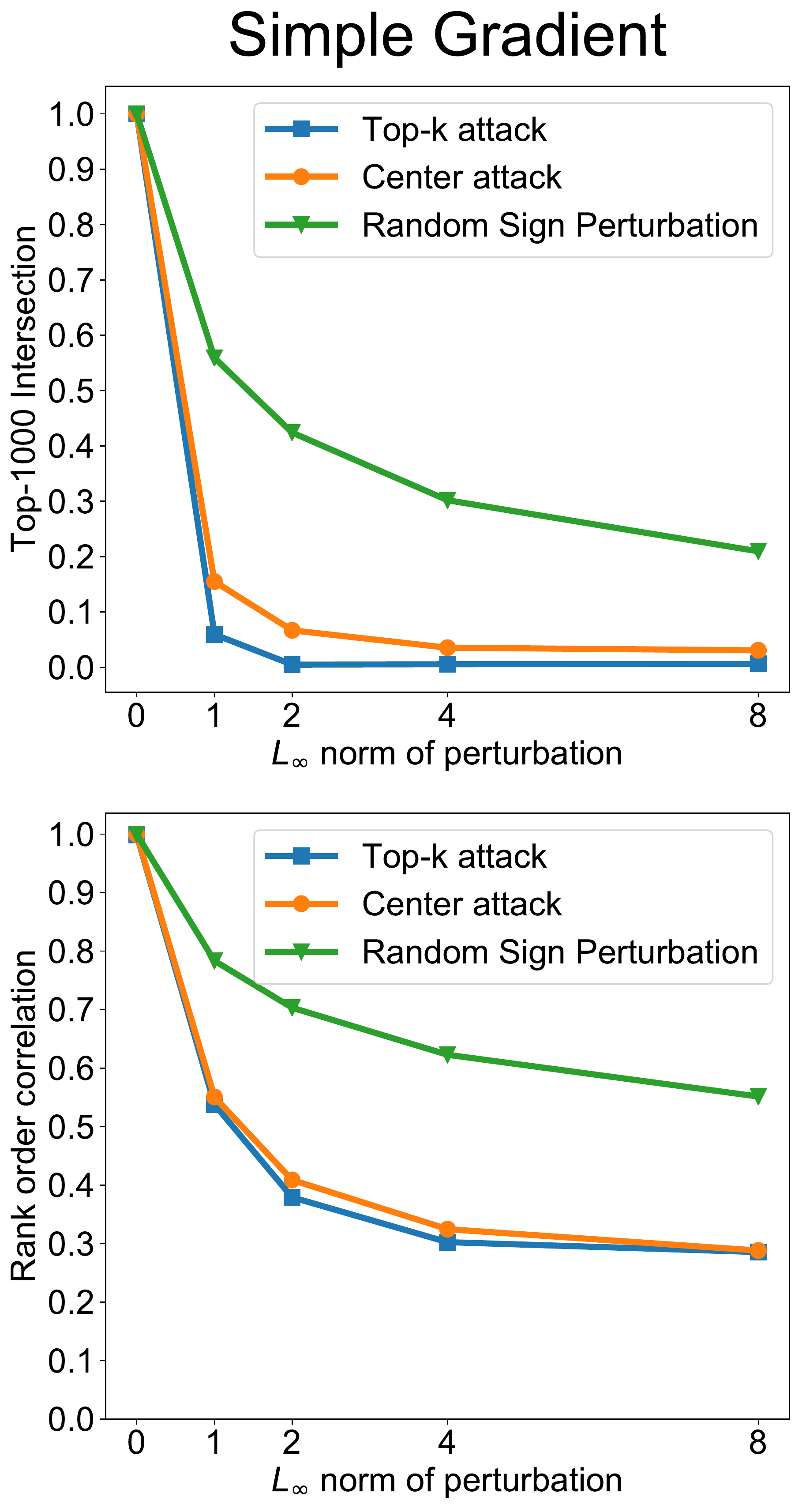}} 
\subfloat[]{\includegraphics[width=0.26\linewidth]{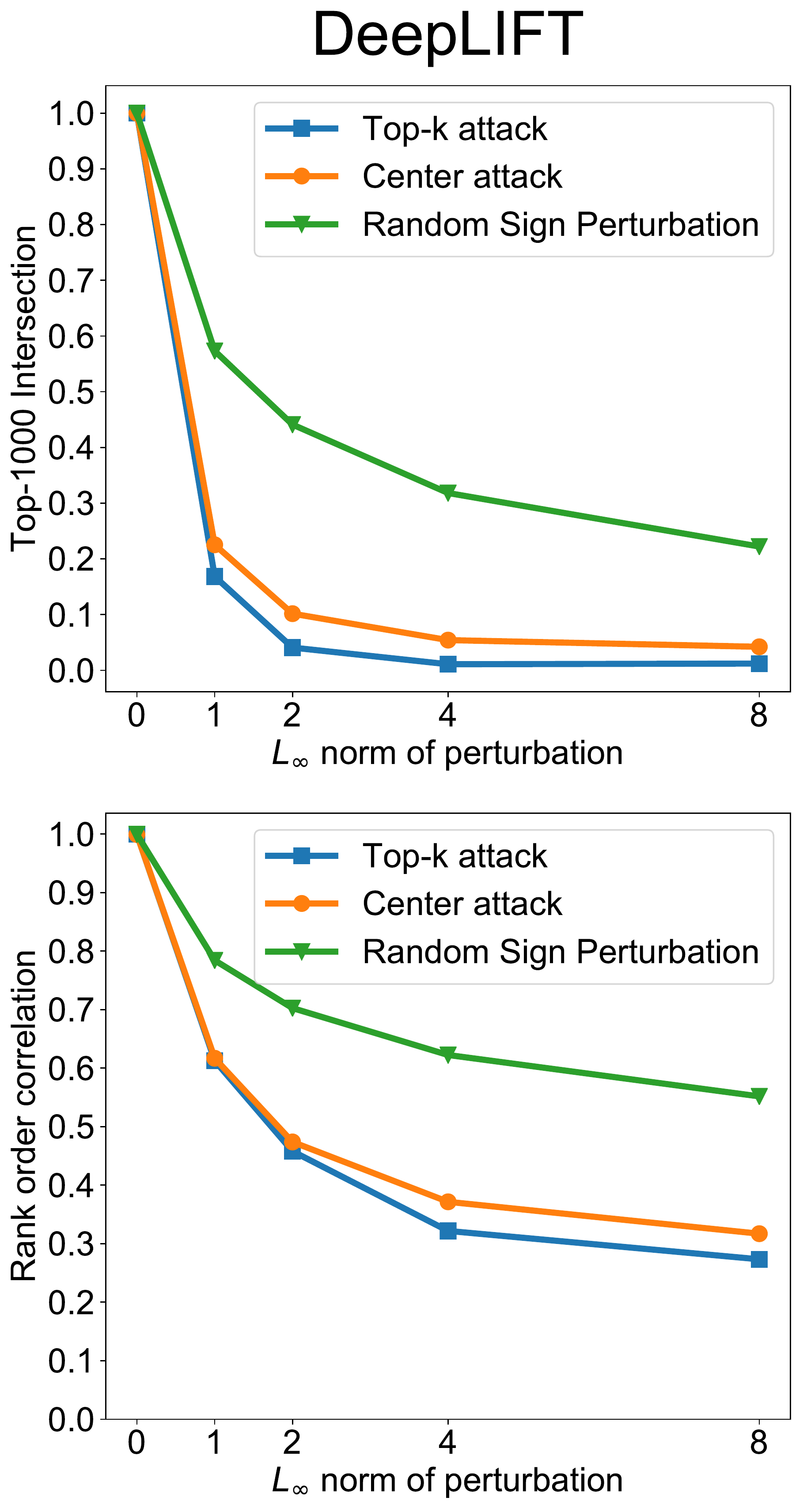}}
\subfloat[]{\includegraphics[width=0.26\linewidth]{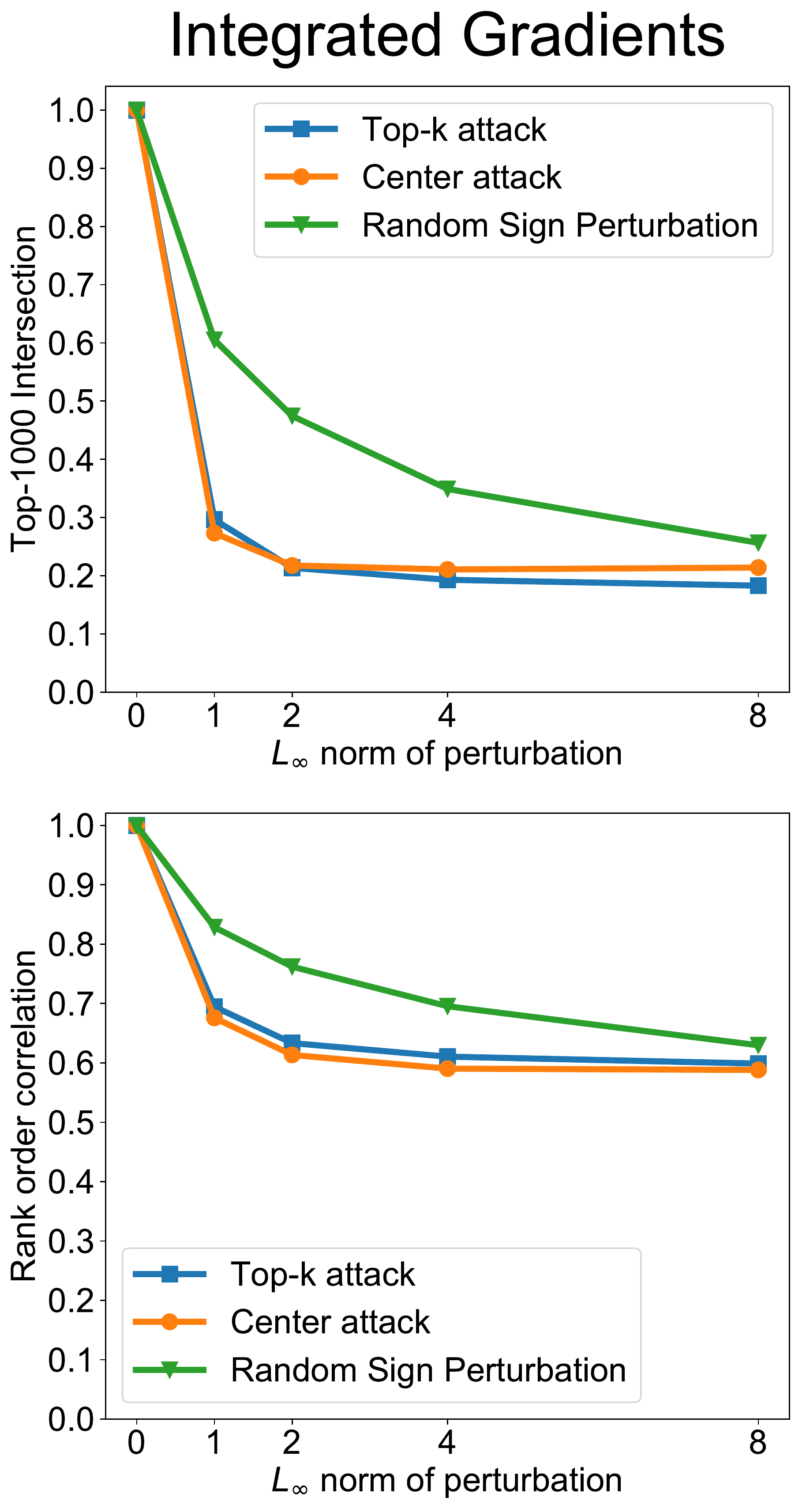}}
\caption{\textbf{Comparison of adversarial attack algorithms on feature-importance methods.} Across 512 correctly-classified ImageNet images, we find that the top-$k$ and center attacks perform similarly in top-1000 intersection and rank correlation measures, and are far more effective than the random sign perturbation at demonstrating the fragility of interpretability, as characterized through top-1000 intersection \textbf{(top)} as well as rank order correlation \textbf{(bottom)}. 
\label{fig:results}
}.
\end{figure*}

In Fig.~\ref{fig:results}, we present results aggregated over all 512 images. We compare different attack methods using top-1000 intersection and rank correlation methods. In all the images, the attacks do not change the original predicted label of the image nor does it significantly change the prediction confidence. 
Random sign perturbation already causes decreases in both top-1000 intersection and rank order correlation. For example, with $L_{\infty} = 8$, on average, there is less than 30\% overlap in the top 1000 most salient pixels between the original and the randomly perturbed images across all three of interpretation methods.

Both the mass-center and top-K attack algorithms have similar effects on feature importance of test images when measured on the basis of rank correlation or top-1000 intersection. We show an additional metric in Appendix \ref{appendix:results} that measures the displacement in feature importance maps and empirically has the most correlation with perceptual change in interpretation. Not surprisingly, we found that the mass-center attack was more effective than the top-$k$ attack at resulting in the most perceptive change. Average numerical results are not obtainable for the targeted attack as it is designed for semantic change and requires a target area of attack in each image. Comparing the effectiveness of attacks among the three different feature importance methods, we found that the integrated gradients method was the most difficult one to generate adversarial examples for. Similar results for CIFAR-10 can be found in Appendix ~\ref{appendix:results_cifar}.

\begin{figure*}[!htb]
\centering
\subfloat[]{\includegraphics[width=0.5\linewidth]{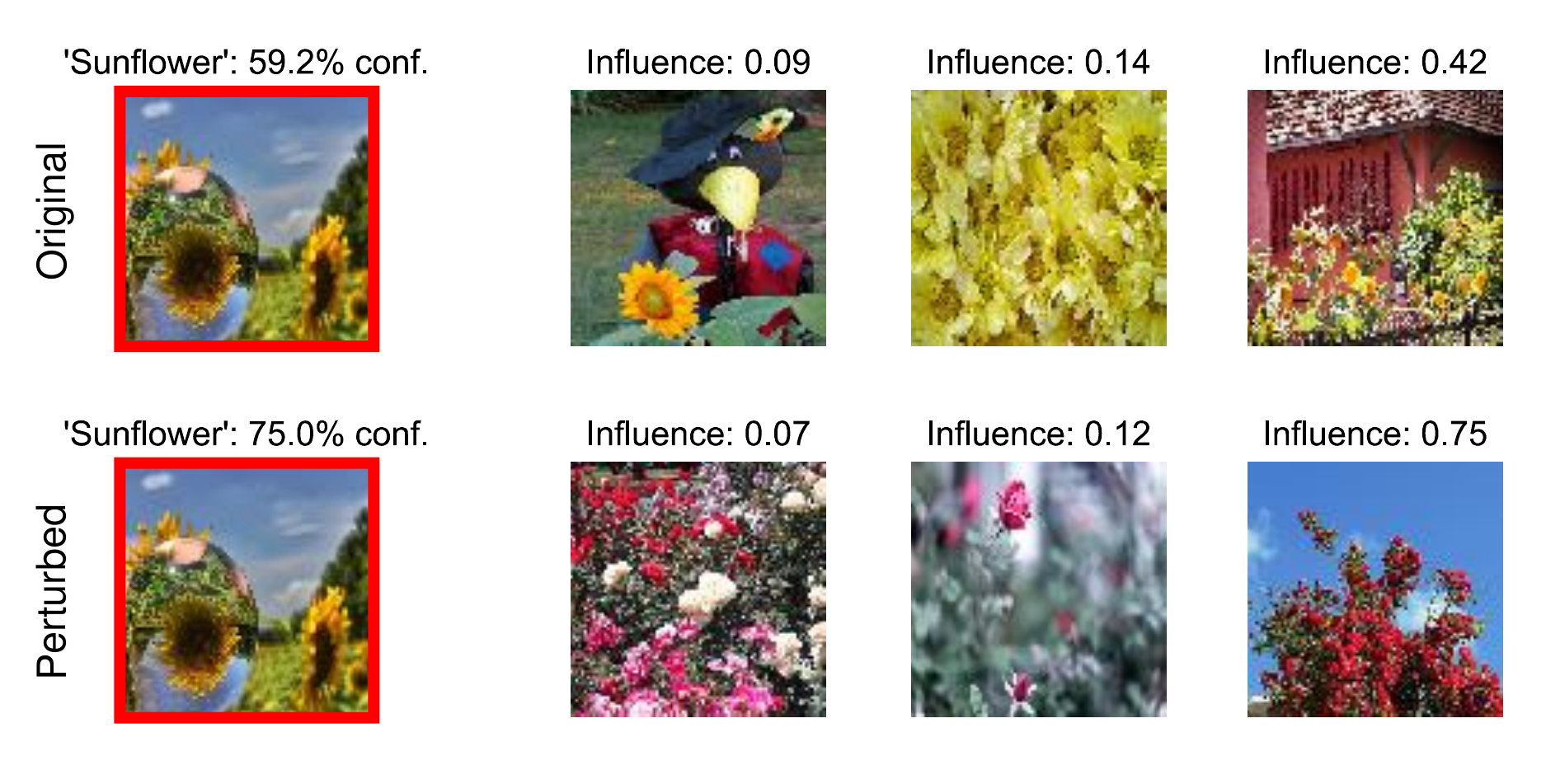}}
\subfloat[]{\includegraphics[width=0.25\linewidth]{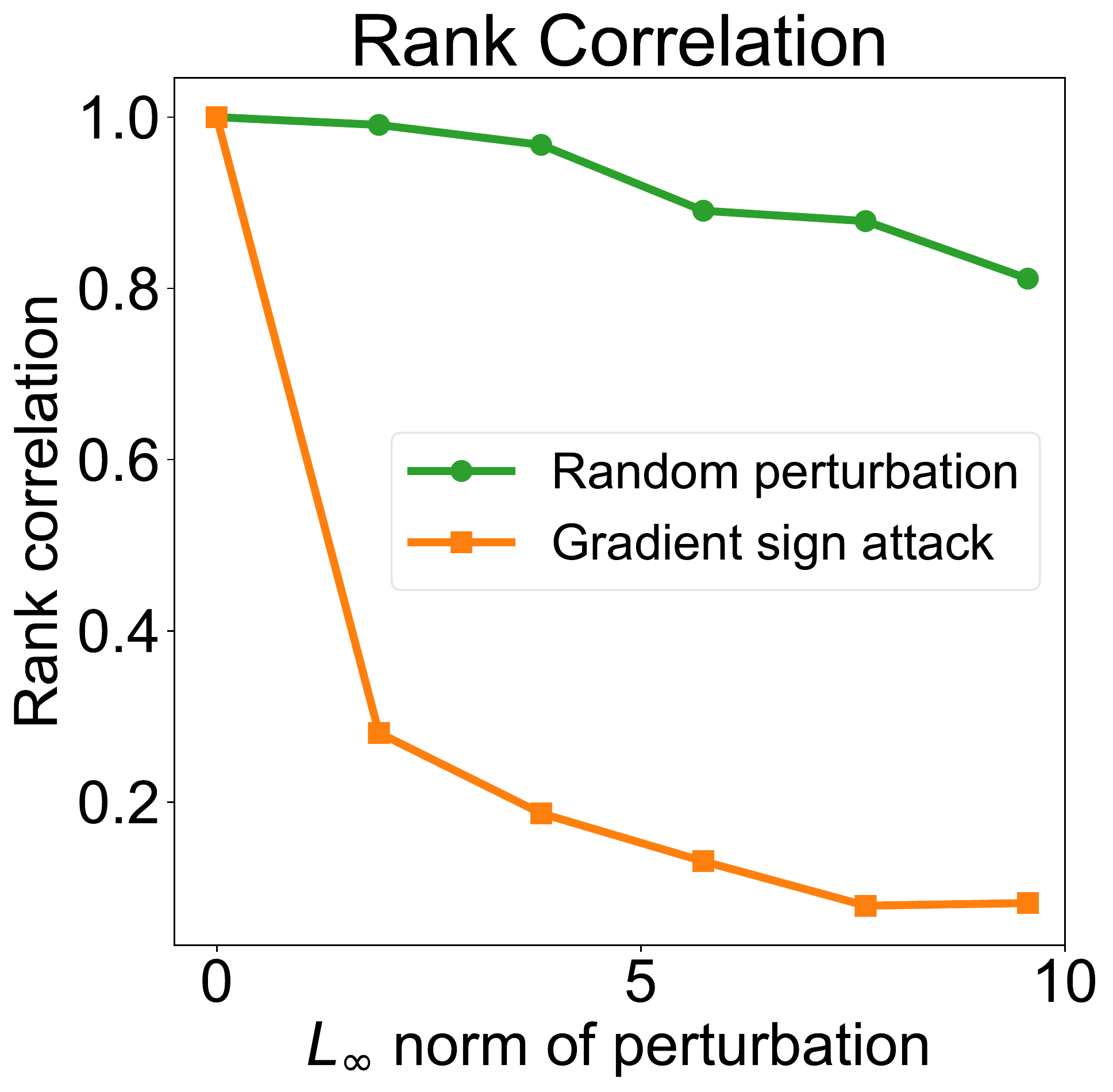}} 
\subfloat[]{\includegraphics[width=0.24\linewidth]{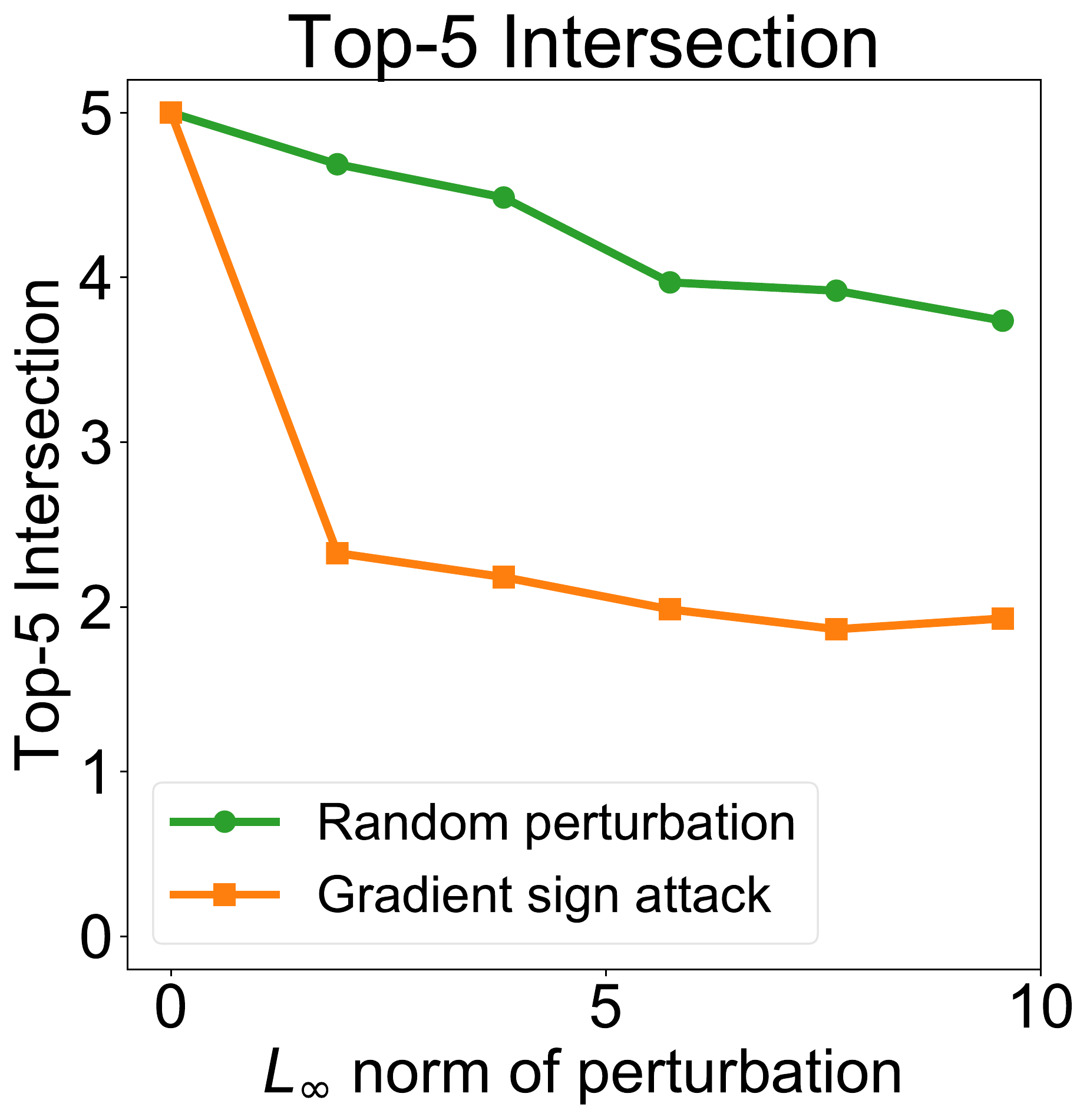}}
\caption{\textbf{Gradient sign attack on influence functions .} (a)  An imperceptible perturbation to a test image can significantly affect sample importance interpretability. The original test image is that of a sunflower that is classified correctly in a rose vs. sunflower classification task. The top 3 training images identified by influence functions are shown in the \textbf{top row}. Using the gradient sign attack, we perturb the test image (with $\epsilon=8$) to produce the leftmost image in the \textbf{bottom row}. Although the image is even more confidently predicted as a sunflower, influence functions suggest very different training images by means of explanation: instead of the sunflowers and yellow petals that resemble the input image, the most influential images are pink/red roses. (b) Average results for applying random (green) and gradient sign-based (orange) perturbations to 200 test images are shown. Random attacks have a gentle effect on interpretability while a gradient perturbation can significantly affect the rank correlation and (c) the 5 most influential images.
Although the image is even more confidently predicted to be a sunflower, influence functions suggest very different training images by means of explanation: instead of the sunflowers and yellow petals that resemble the input image, the most influential images are pink/red roses. 
The plot on the right shows the influence of each training image before and after perturbation. The 3 most influential images (targeted by the attack) have decreased in influence, but the influences of other images have also changed.
}
\label{fig:gradient-unstable1}
\end{figure*}

\paragraph{Results for adversarial attacks against sample importance scores.}
We evaluate the robustness of influence functions on a test data set consisting of 200 images of roses and sunflowers. Fig. \ref{fig:gradient-unstable1}(a) shows a representative test image to which we have applied the gradient sign attack. Although the prediction of the image does not change, the most influential training examples change entirely. Additional examples can be found in Appendix \ref{appendix:more_influence_examples}.

In Fig. \ref{fig:gradient-unstable1}(b,c), we compare the random perturbations and gradient sign attacks for the test set. It shows that gradient sign-based attacks are significantly more effective at decreasing the rank correlation, as well as distorting the top-5 influential images. For example, on average, with a  perturbation of magnitude $\epsilon=8$, only 2 of the top 5 most influential training images remain in the top 5. The influences of the training images before and after an adversarial attack are essentially uncorrelated. However, we find that even random attacks can have a small but non-negligible effect on influence functions, on average reducing the rank correlation to 0.8 ($\epsilon \approx 10$). 

\begin{figure*}[ht]
\centering
\includegraphics[width=\linewidth]{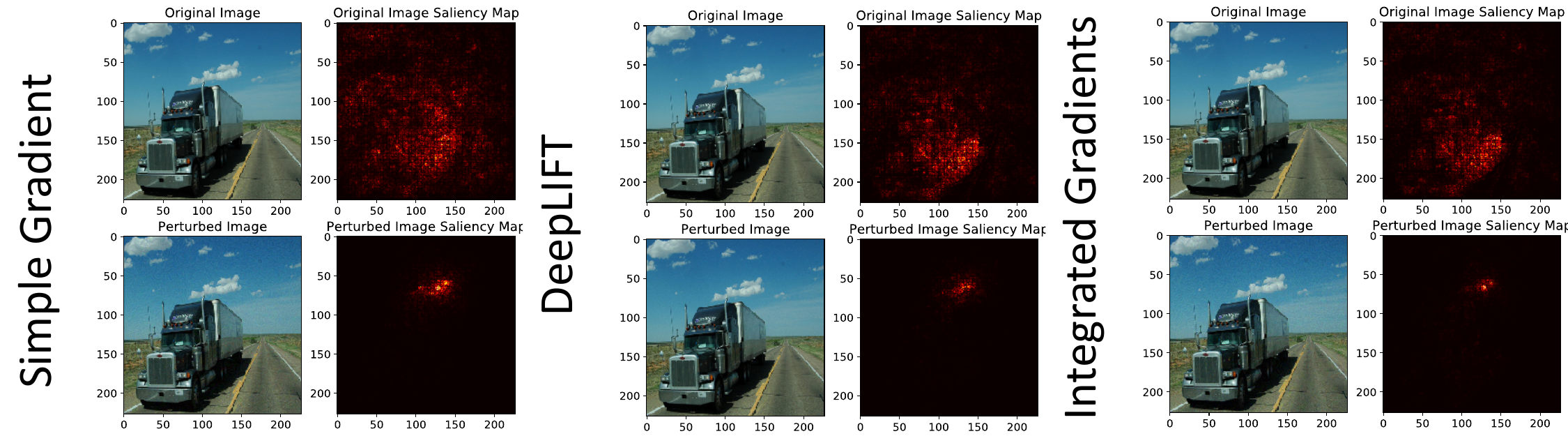}
\caption{\textbf{Targeted attack against feature importance map.} Image is correctly classified as “trailer truck”. For all methods, the devised perturbation with $\epsilon=8$ was able to semantically meaningfully change the focus of saliency map to the “cloud” above the truck. (The cloud area was captured using SLIC~\protect\cite{achanta2012slic} superpixel segementation.)
 \textbf{(top)} as well as rank order correlation \textbf{(bottom)}. \label{fig:semantic}}
\end{figure*}

\section{Hessian Analysis}
\label{section:geometric}

In this section, we explain the effectiveness of adversarial attacks on interpretations in terms of the high dimensionality and non-linearities in deep networks. High dimensionality is also a reason why adversarial examples are effective at changing prediction labels \cite{goodfellow2014explaining}.

Let $S(\xb; \wb)$ denote the score function of interest and $\xb \in \mathbb{R}^d$ be the input vector. First order approximation of sensitivity of a gradient-based interpretation to perturbations in the input is  $\boldsymbol{\delta} \in \mathbb{R}^d$ is: 
$\nabla_{\xb} S(\xb+\deltb) - \nabla_{\xb}S(\xb) \approx H \deltb $, where $H$ is the Hessian matrix $H_{i,j} = \frac{\partial S}{\partial x_i \partial x_j}$. In the most simple case of having a linear model $S = \wb^{\top} \xb$, the feature importance vector is robust as it is completely independent of $\xb$ ($\nabla_{\xb}S = \wb$). Thus, some non-linearity is required for adversarial attacks against interpretation. The simplest model susceptible to interpretation adversarial attacks is a set of weights followed by a non-linearity (e.g. softmax): $S = g(\boldsymbol{w}^\top \boldsymbol{x})$.

The first order approximation of change in feature importance map due to a small input perturbation: $\boldsymbol{x} \rightarrow \boldsymbol{x} + \boldsymbol{\delta}$ will be equal to : $H \cdot  \boldsymbol{\delta}=\nabla^2_{\boldsymbol{x}} S \cdot  \boldsymbol{\delta}$. In particular, the relative change in the importance score of the $i^{\text{th}}$ feature is $(\nabla^2_{\boldsymbol{x}} S \cdot  \boldsymbol{\delta})_i/(\nabla_{\boldsymbol{x}} S )_i$. For our simple model, this relative change is:
\begin{align}
\frac{(\boldsymbol{w}\boldsymbol{w}^{\top}\boldsymbol{\delta}g''(\boldsymbol{w}^{\top}\boldsymbol{x}))_i}{(\boldsymbol{w}g'(\boldsymbol{w}^{\top}\boldsymbol{x}))_i} = 
\frac{ \boldsymbol{w}^{\top}\boldsymbol{\delta}g''(\boldsymbol{w}^{\top} \boldsymbol{x})}{g'(\boldsymbol{w}^{\top}\boldsymbol{x})},
\end{align}
where we have used $g'(\cdot)$ and $g''(\cdot)$ to refer to the first and second derivatives of $g(\cdot)$. Note that $g'(\boldsymbol{w}^{\top}\boldsymbol{x})$ and $g''(\boldsymbol{w}^{\top}\boldsymbol{x})$ do not scale with the dimensionality of $\boldsymbol{x}$ because in general, $\boldsymbol{x}$ and $\boldsymbol{w}$ are $\ell_2$-normalized or have fixed $\ell_2$-norm due to data preprocessing and weight decay regularization. However, if $\boldsymbol{\delta} = \epsilon\text{sign}(\boldsymbol{w})$, then the relative change in the feature importance grows with the dimension, since it is proportional to the $\ell_1$-norm of $\wb$. For a high dimensional input the relative effect of the perturbation can be substantial.  Note also that this perturbation is exactly the sign of the first right singular vector of the Hessian $\nabla^2_{\boldsymbol{x}} S$, which is appropriate since that is the vector that has the maximum effect on the gradient of $S$. (Similar analysis for influence functions in Appendix \ref{appendix:influence_dimensionality_analysis}). 

\begin{figure}[!htb]
\centering

{\caption{\textbf{Orthogonality of Prediction and Interpretation Fragile Directions} (a) The histogram of the angle between the steepest direction of change in (a) feature importance and (b) samples importance and the steepest prediction score change direction\label{fig:angle_hist}.}}
{
\subfloat[]{\includegraphics[width=0.51\linewidth]{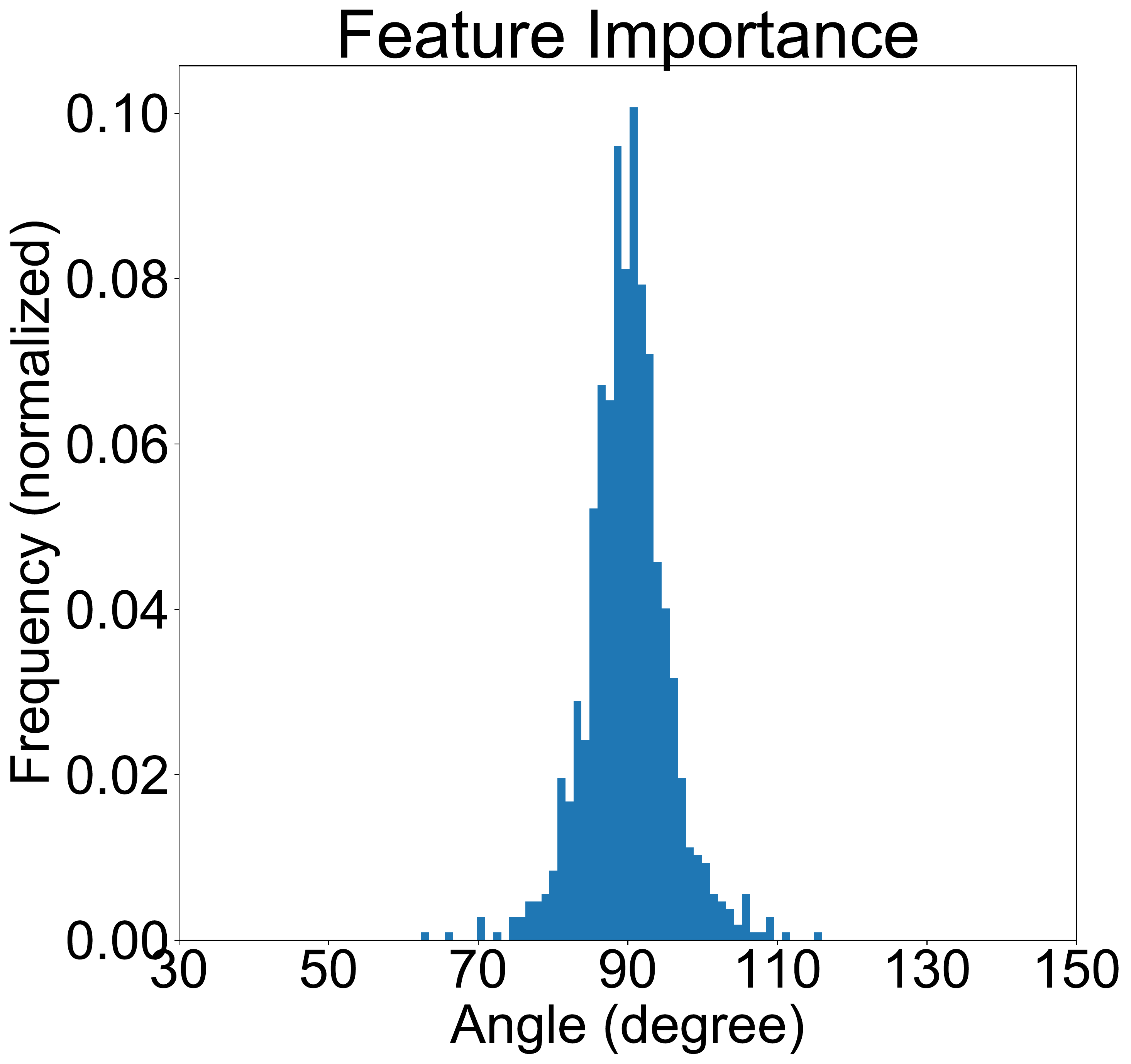}}
\subfloat[]{\includegraphics[width=0.48\linewidth]{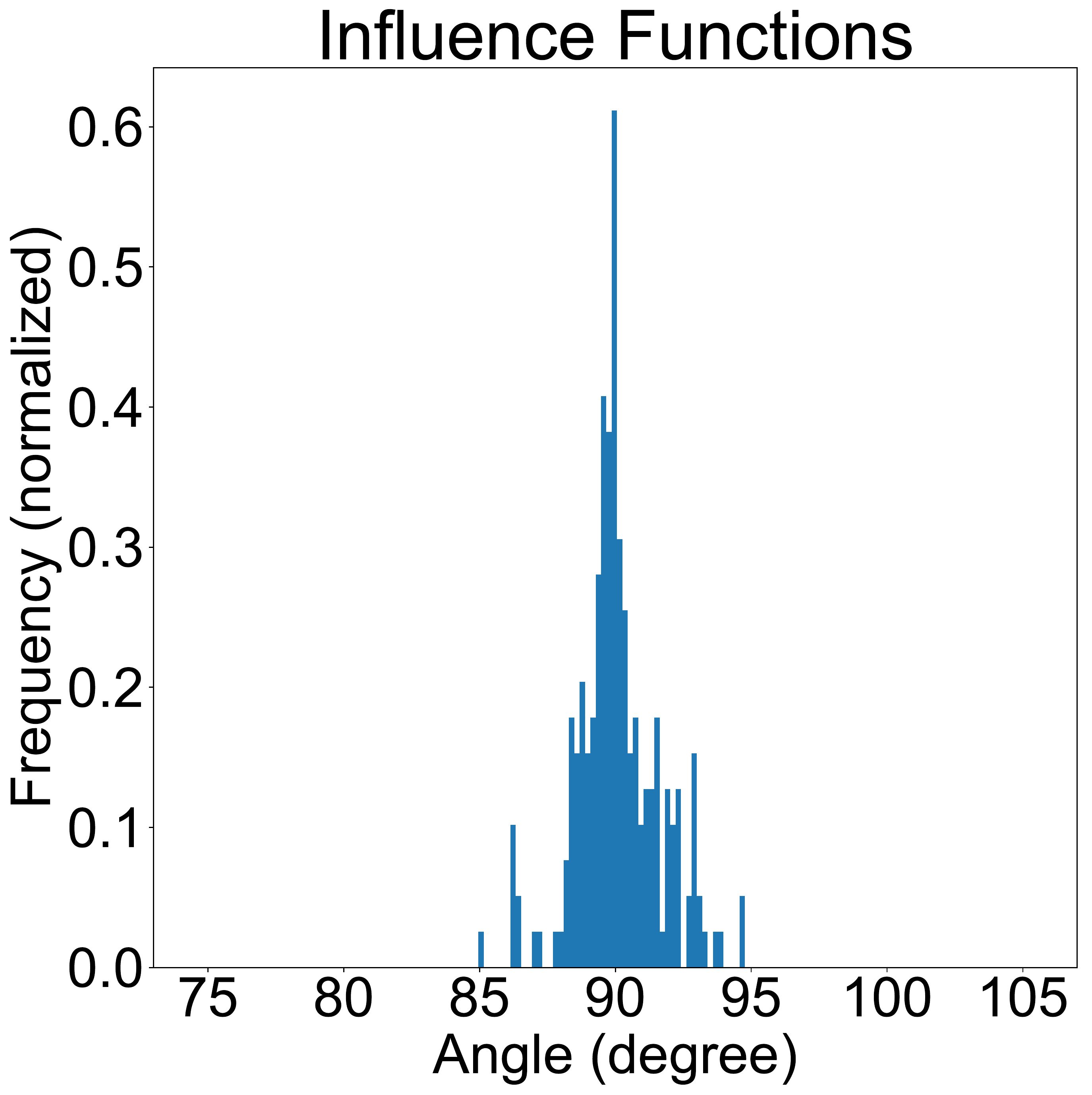}}
}

\end{figure}

Notice that for this simple network, the direction of adversarial attack on interpretability, $\text{sign}(\wb)$ is the same as the adversarial attack on prediction which means that perturbing interpretability perturbs prediction. For more complex networks, this is not the case and in Appendix~\ref{appendix:two-layer} we show this analytically for a simple case of a two-layer network. As an empirical test, in Fig.~\ref{fig:angle_hist}(a), we plot the distribution of the angle between most fragile directions of interpretation and prediction for 1,000 CIFAR-10 images (Model details in Appendix~\ref{appendix:CIFAR10-network}). Fig.~\ref{fig:angle_hist}(b) shows the equivalent distribution for influence functions, computed across all 200 test images.

\section{Discussion}



This paper demonstrates that interpretation of neural networks can be fragile in the sense that two similar inputs with the same predictedion can be given very different interpretations. We develop perturbations to illustrate this fragility and propose evaluation metrics as well as insights on why fragility occurs. Fragility of neural network interpretation can be orthogonal to fragility of the prediction, as we demonstrate with perturbations that substantially change the interpretation without changing the predicted label, but both types of fragility arise at least in part from high dimensionality, as we discuss in Section~\ref{section:geometric}. 

Our main message is that robustness of the interpretation of a prediction is an important and challenging problem, especially as in many applications (e.g. many biomedical and financial settings), users are as interested in the interpretation as in the prediction itself. Our results raise concerns on how interpretations of neural networks can be manipulated. Especially in settings where the importance of individual or a small subset of features are interpreted, we show that these importance scores can be sensitive to even random perturbation. More dramatic manipulations of interpretations can be achieved with our targeted perturbations. This is especially true for the simple gradients method, DeepLIFT, and influence functions, but also the integrated gradients method. These results raise potential security concerns. We do not suggest that interpretations are meaningless, just as adversarial attacks on predictions do not imply that neural networks are useless. Interpretation methods do need to be used and evaluated with caution while applied to neural networks, as they can be fooled into identifying features that would not be considered salient by human perception. 

Our results demonstrate that the \emph{interpretations} (e.g. saliency maps) are vulnerable to perturbations, but this does not imply that the \emph{interpretation methods} are broken by the perturbations. This is a subtle but important distinction. Methods such as saliency measure the infinitesimal sensitivity of the neural network at a particular input $\xb$. After a perturbation, the input has changed to $\tilde{\xb} = \xb + \deltb$, and the saliency now measures the  sensitivity at the perturbed input. The saliency \emph{correctly} captures the infinitesimal sensitivity at the two inputs; it's doing what it is supposed to do. The fact that the two resulting saliency maps are very different is fundamentally due to the network itself being fragile to such perturbations, as we illustrate with Fig.~\ref{fig:concept}. 

Our work naturally raises the question of how to defend against adversarial attacks on interpretation. Because interpretation fragility arises as a consequence of high dimensionality and non-linearity (see section \ref{section:geometric}), we believe that techniques that discretize inputs, such as thermometer encoding \cite{buckman2018thermometer}, and train neural networks in a way to constrain the non-linearity of the network \cite{moustapha2017parseval}, may be useful in defending against interpretation attacks. 

While we focus on  the standard image benchmarks for popular interpretation tools, this fragility issue can be wide-spread in biomedical, economic and other settings where neural networks are increasingly used. Understanding interpretation fragility in these applications and developing more robust methods are important agendas of research.

\newpage\phantom{blabla}
\newpage\phantom{blabla}

\empty

\bibliography{main}
\bibliographystyle{aaai}

\appendix
\newpage
\begin{onecolumn}

\section*{Appendices}
\section{Description of the CIFAR-10 classification network}
\label{appendix:CIFAR10-network}

We trained the following structure using ADAM optimizer ~\cite{kingma2014adam} with default parameters. The resulting test accuracy using ReLU activation was 73\%. For the experiment in Fig,~\ref{fig:angle_hist}(a), we replaced ReLU activation with Softplus and retrained the network (with the ReLU network weights as initial weights). The resulting accuracy was 73\%.
\begin{center}
 \begin{tabular}{||c||} 
 \hline
 Network Layers \\ [0.5ex] 
 \hline\hline
 $3 \times 3$ conv. 96 ReLU\\
 \hline
$3 \times 3$  conv. 96 ReLU\\
 \hline
$3 \times 3$  conv. 96 Relu\\
 Stride 2\\
 \hline
 $3 \times 3$ conv. 192 ReLU\\
 \hline
$3 \times 3$  conv. 192 ReLU\\
 \hline
$3 \times 3$  conv. 192 Relu\\
Stride 2\\
 \hline
 1024 hidden sized feed forward\\
 \hline \hline
 \end{tabular}
\end{center}

\newpage
\section{Additional examples of feature importance perturbations}
Here, we provide three more examples from ImageNet. For each example, all three methods of feature importance  are attacked by random sign noise and our two adversarial algorithms.
\label{appendix:examples}

\begin{figure}[H]
\centering
\includegraphics[width=0.9\linewidth]{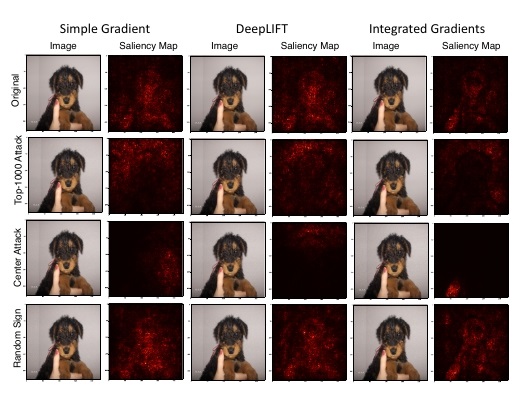}
\caption{ All of the images are classified as a \emph{airedale}.}
\end{figure}
\newpage

\begin{figure}[H]
\centering
\includegraphics[width=0.9\linewidth]{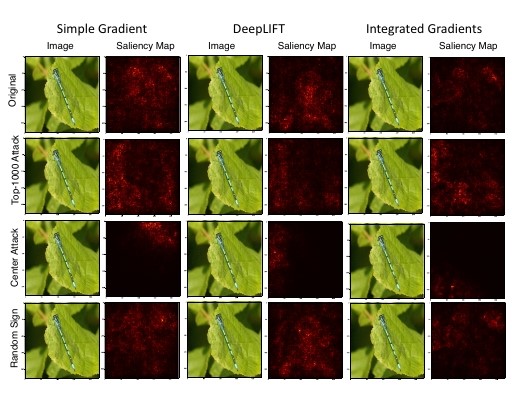}
\caption{  All of the images are classified as a \emph{damselfly}.}
\end{figure}
\newpage

\begin{figure}[H]
\centering
\includegraphics[width=0.9\linewidth]{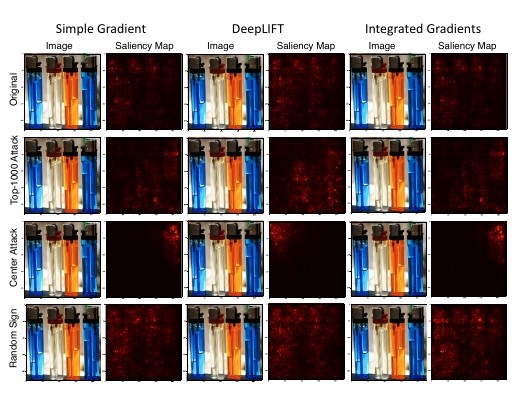}
\caption{ All of the images are classified as a \emph{lighter}.}
\end{figure}

\section{Objective metrics and subjective change in feature importance maps}
\label{appendix:metrics}
\begin{figure}[H]
\centering
\includegraphics[width=0.65\linewidth]{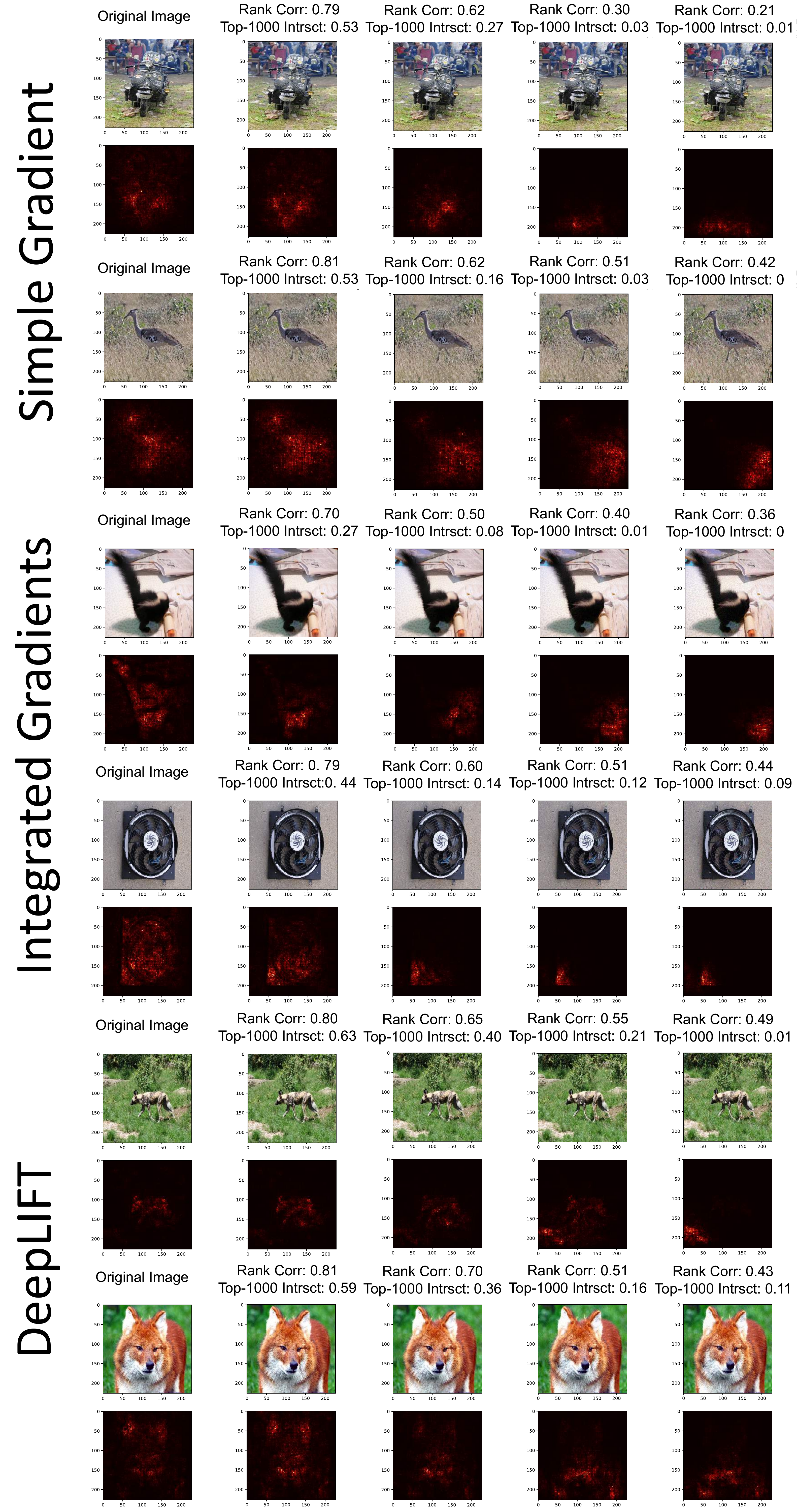}
\caption{\textbf{Evaluation metrics vs subjective change in feature importance maps}
To have a better sense of how rank order correlation and top-1000 intersection metrics are related to changes in feature importance maps, snapshots of the iterations of mass-center attack are depicted.}

\label{fig:metr}
\end{figure}

\section{Semantically meaningful change in feature importance using targeted attack}
\label{appendix:semantic}
\begin{figure}[H]
\centering
\includegraphics[width=0.4\linewidth]{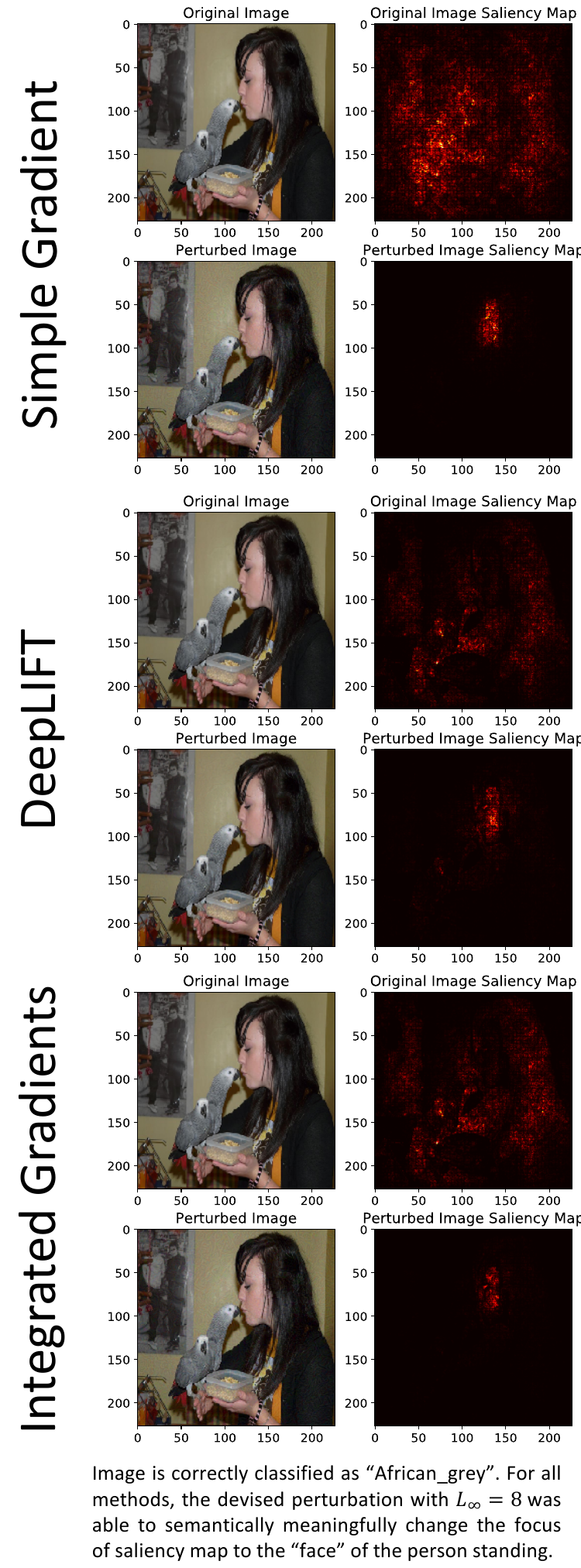}
\caption{\textbf{Semantically meaningful interpretation perturbation}
Using targeted attack method, we can change the the most salient object in the image to be different while not affecting the predicted label.}

\label{fig:metr}
\end{figure}

\section{Measuring center of mass movement}
\label{appendix:results}
\begin{figure}[H]
\centering
\includegraphics[width=\linewidth]{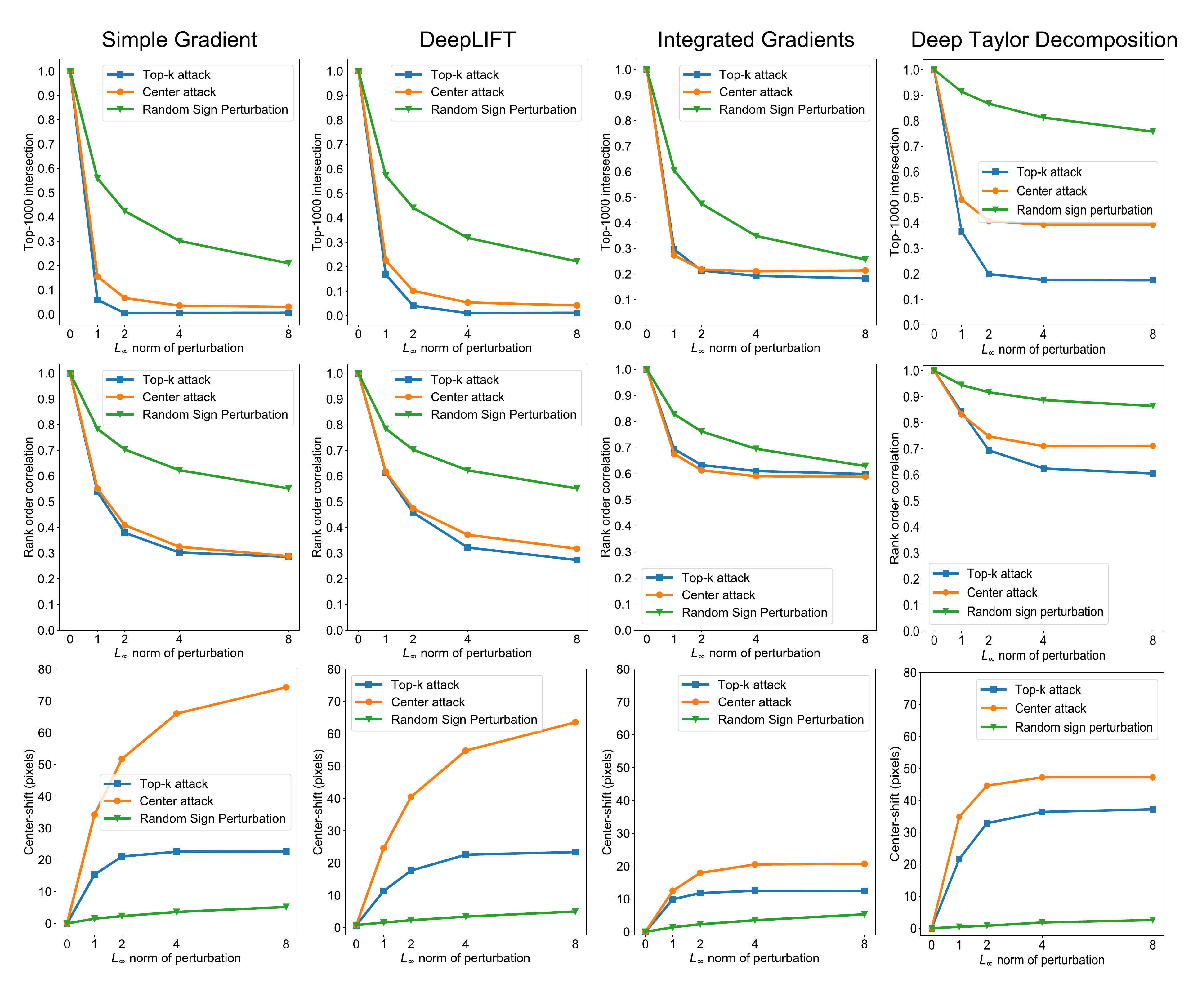} 

\caption{\textbf{Aggregate results for four feature importance methods on ImageNet}: As discussed in the paper, among our three measurements, center-shift measure was the most correlated measure with the subjective perception of change in feature importance maps. The results in Appendix~\ref{appendix:examples} also show that the center attack which resulted in largest average center-shift, also results in the most significant subjective change in feature importance maps. Random sign perturbations, on the other side, did not substantially change the global shape of the feature importance maps, though local pockets of feature importance are sensitive. Just like rank correlation and top-1000 intersection measures, the integrated gradients method is the most robust method against adversarial attacks in the center-shift measure .
\label{fig:examples2}}
\end{figure}
\newpage

\section{Results for adversarial attacks against CIFAR-10 feature importance methods}
\label{appendix:results_cifar}

\begin{figure}[H]
\centering
\subfloat[]{\includegraphics[width=0.33\linewidth]{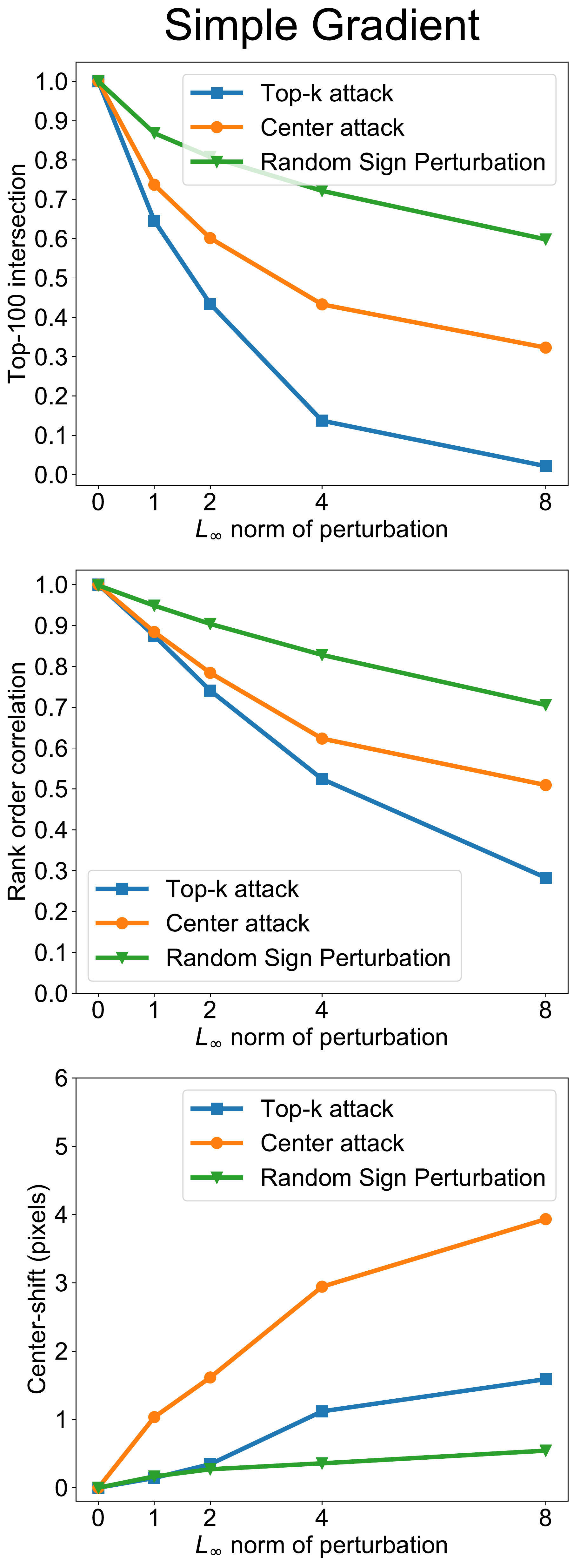}} 
\subfloat[]{\includegraphics[width=0.33\linewidth]{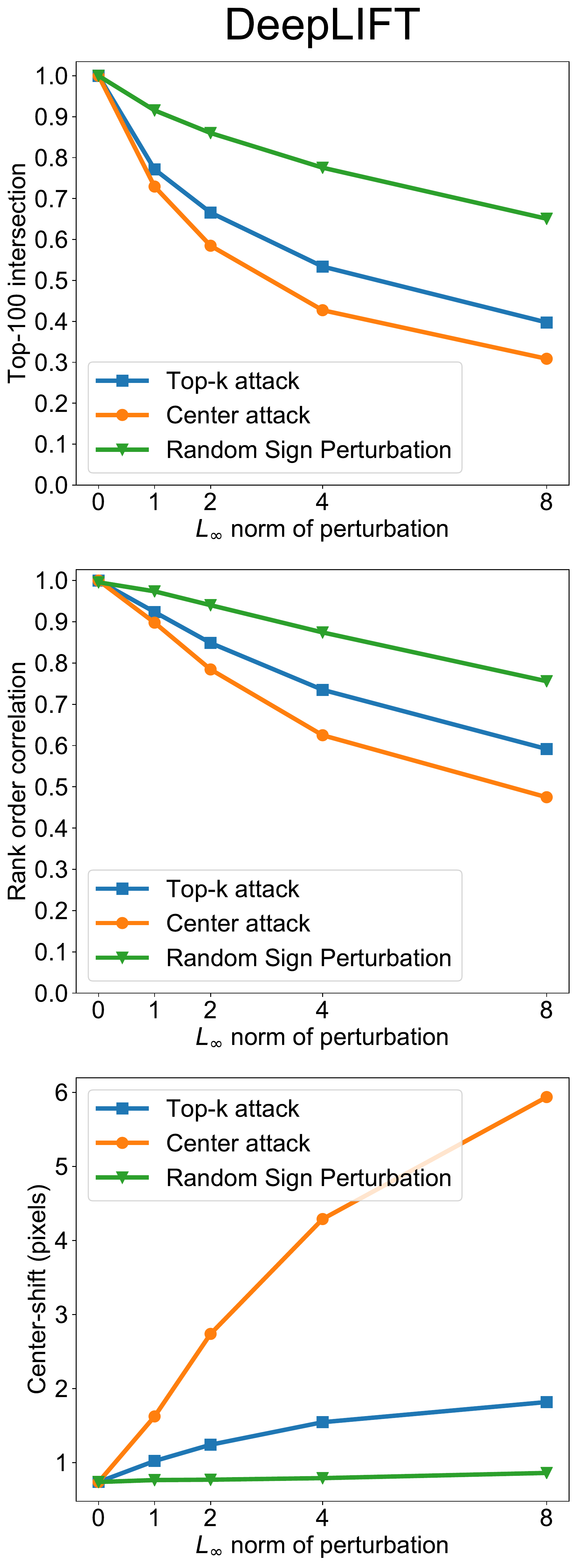}}
\subfloat[]{\includegraphics[width=0.33\linewidth]{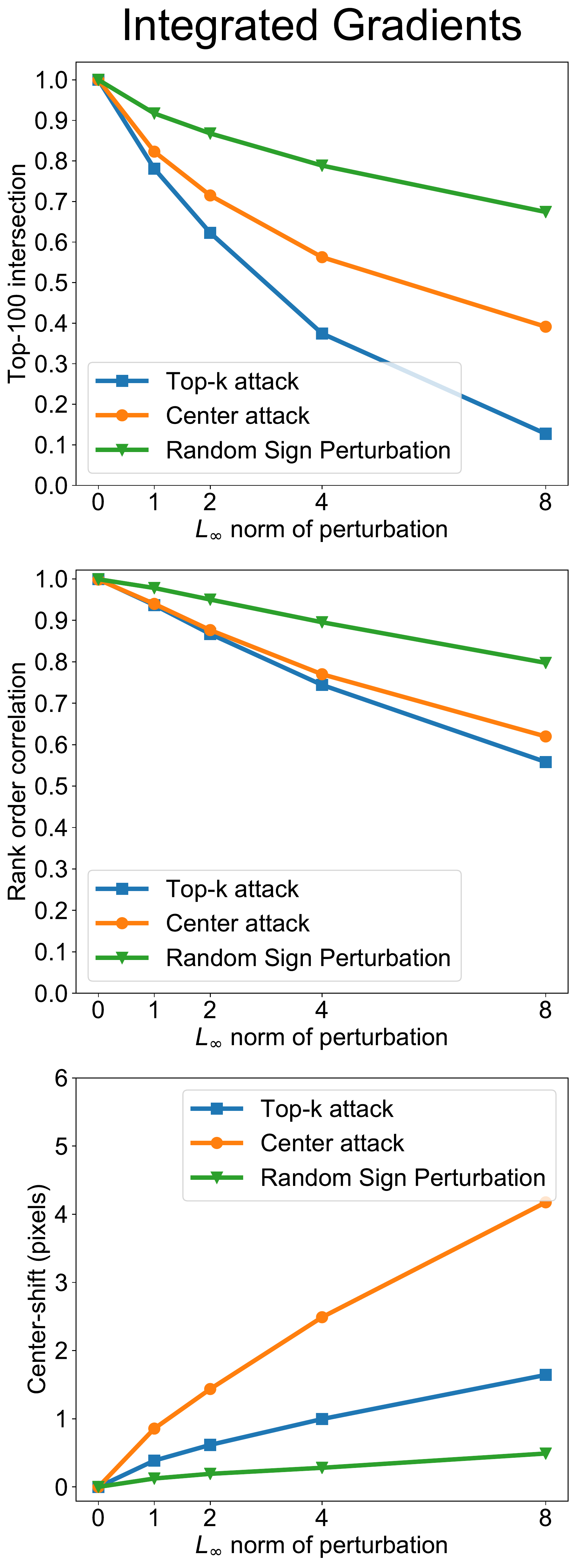}}
\caption{\textbf{Results for adversarial attacks against CIFAR10 feature importance methods}:  
\label{fig:cifar10_results} For CIFAR10 the mass-center attack and top-k attack with k=100 achieve similar results for rank correlation and top-100 intersection measurements and both are stronger than random perturbations. Mass-center attack moves the center of mass more than two other perturbations. Among different feature importance methods, integrated gradients is more robust than  the two other methods. Additionally, results for CIFAR10 show that images in this data set are more robust against adversarial attack compared to ImageNet images which agrees with our analysis that higher dimensional inputs are tend to be more fragile.}
\end{figure}

\newpage
\section{Additional examples of adversarial attacks on influence functions}
\label{appendix:more_influence_examples}

In this appendix, we provide additional examples of the fragility of influence functions, analogous to Fig. \ref{fig:gradient-unstable1}. 

\begin{figure}[H]
\centering
\subfloat[]{\includegraphics[width=\linewidth]{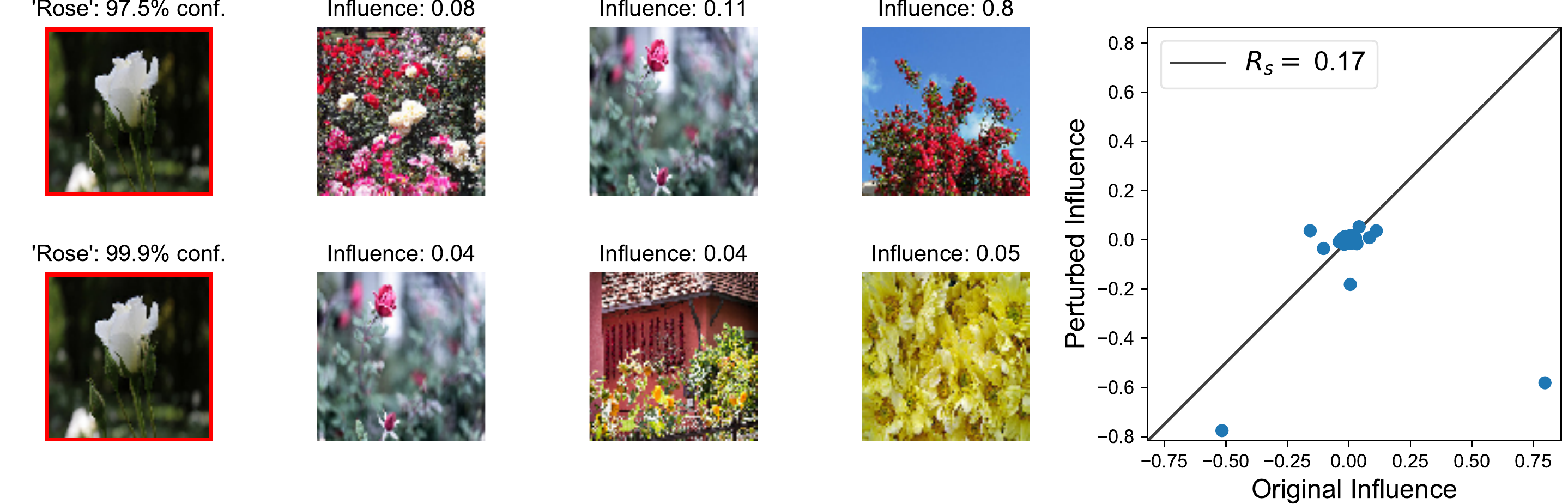}} \\
\subfloat[]{\includegraphics[width=\linewidth]{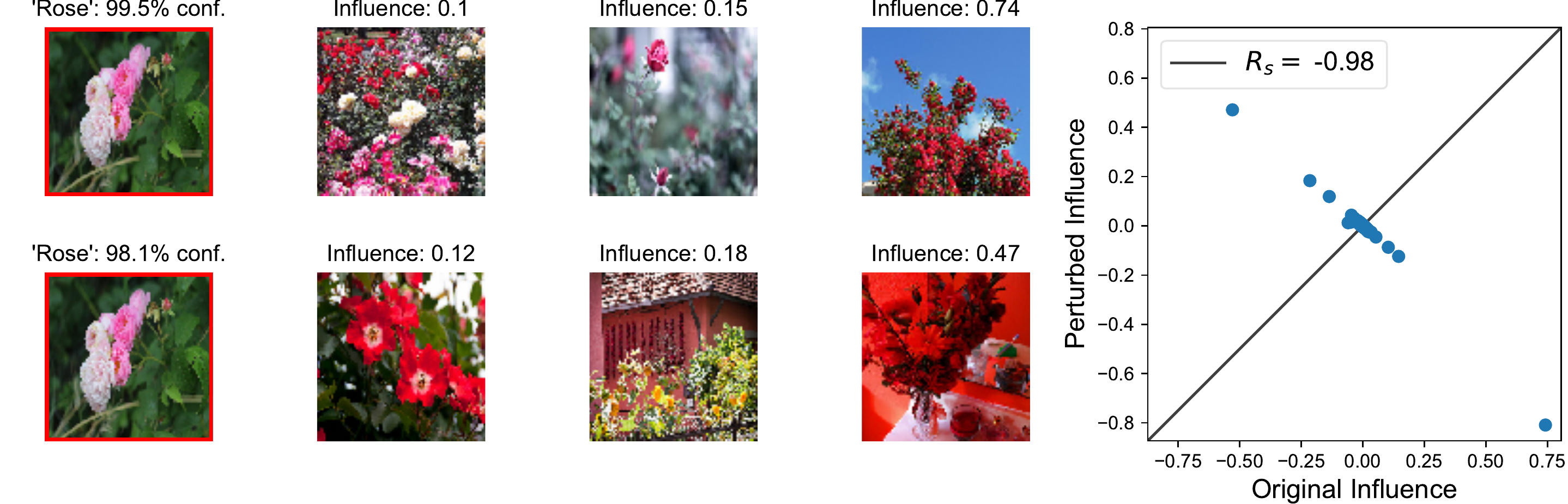}}
\caption{\textbf{Further examples of gradient-sign attacks on influence functions.} (a) Here we see a  representative example of the most influential training images before and after a perturbation to the test image. The most influential image before the attack is one of the least influential afterwards. Overall, the influences of the training images before and after the attack are uncorrelated. 
(b) In this example, the perturbation has remarkably caused the training images to almost completely reverse in influence. Training images that had the most positive effect on prediction now have the most negative effects and the other way round.}
\end{figure}

\newpage
\section{Dimensionality-based explanation for fragility of influence functions}
\label{appendix:influence_dimensionality_analysis}

Here, we demonstrate that increasing the dimension of the input of a simple neural network increases the fragility of that network with respect to influence functions, analogous to the calculations carried out for importance-feature methods in Section \ref{section:geometric}. Recall that the influence of a training image $z_i = (\xb_i, y_i)$ on a test image $z = (\xb, y)$ is given by:
\begin{align}
I(z_i, z) = - \underbrace{\nabla_\theta L(z, \hat{\theta})^\top}_{\text{dependent on}\, \xb} \underbrace{H_{\hat{\theta}}^{-1} \nabla_\theta L(z_i, \hat{\theta})}_{\text{independent of}\, \xb}.
\label{eqn:app_def_infl}
\end{align}
We restrict our attention to the term in (\ref{eqn:app_def_infl}) that is dependent on $\xb$, and denote it by $J = \nabla_\theta L$. $J$ represents the infinitesimal effect of each of the parameters in the network on the loss function evaluated at the test image. 

Now, let us calculate the change in this term due to a small perturbation in $\boldsymbol{x} \rightarrow \boldsymbol{x} + \boldsymbol{\delta}$. The first-order approximation for the change in $J$ is equal to: $\nabla_{\xb} J \cdot  \boldsymbol{\delta}=\nabla_\theta \nabla_{\boldsymbol{x}} L \cdot  \boldsymbol{\delta}$. In particular, for the $i^\text{th}$ parameter, $J_i$ changes by $(\nabla_\theta \nabla_{\boldsymbol{x}} L \cdot  \boldsymbol{\delta})_i$ and furthermore, the relative change is $(\nabla_\theta \nabla_{\boldsymbol{x}} L \cdot  \boldsymbol{\delta})_i/(\nabla_{\theta} L )_i$ . For the simple network defined in Section \ref{section:geometric}, this evaluates to (replacing $\theta$ with $\wb$ for consistency of notation):

\begin{align}
\frac{(\boldsymbol{x}\boldsymbol{w}^{\top}\boldsymbol{\delta}g''(\boldsymbol{w}^{\top}\boldsymbol{x}))_i}{(\boldsymbol{x}g'(\boldsymbol{w}^{\top}\boldsymbol{x}))_i} = \frac{x_i \boldsymbol{w}^{\top}\boldsymbol{\delta} g''(\boldsymbol{w}^{\top} \boldsymbol{x})}{x_i g'(\boldsymbol{w}^{\top}\boldsymbol{x})} = \frac{ \boldsymbol{w}^{\top}\boldsymbol{\delta} g''(\boldsymbol{w}^{\top} \boldsymbol{x})}{g'(\boldsymbol{w}^{\top}\boldsymbol{x})},
\end{align}

where for simplicity, we have taken the loss to be $L = |y - g(\wb^{\top}x)|$, making the derivatives easier to calculate. Furthermore, we have used $g'(\cdot)$ and $g''(\cdot)$ to refer to the first and second derivatives of $g(\cdot)$. Note that $g'(\boldsymbol{w}^{\top}\boldsymbol{x})$ and $g''(\boldsymbol{w}^{\top}\boldsymbol{x})$ do not scale with the dimensionality of $\boldsymbol{x}$ because $\boldsymbol{x}$ and $\boldsymbol{w}$ are generalized $L_2$-normalized due to data preprocessing and weight decay regularization. 

However, if we choose $\boldsymbol{\delta} = \epsilon\text{sign}(\boldsymbol{w})$, then the relative change in the feature importance grows with the dimension, since it is proportional to the $L_1$-norm of $\wb$.

\newpage
\section{Orthogonality of steepest directions of change in score and feature importance functions in a Simple Two-layer network}
\label{appendix:two-layer}

Consider a two layer neural network with activation function $g(\cdot)$, input $\xb \in \mathbb{R}^d$, hidden vector  $\ub \in \mathbb{R}^h$ , and score function $S)$, we have:
$$
S = \vb \cdot \ub = \sum_{j=1}^h v_j u_j
$$
$$
\ub = g(W^T \xb) \rightarrow u_j = \wb_j . \xb
$$
where $\wb_j = ||\wb_j||_2 \hat{\wb}_j$. We have:
$$
\nabla_{\xb} S = \sum_{j=1}^h v_j \nabla_{\xb} u_j = \sum_{j=1}^h v_j g^{'} (\wb_j.\xb) \wb_j 
$$
$$
\nabla^2_{\xb} S = \sum_{j=1}^h v_j \nabla^2_{\xb} u_j = \sum_{j=1}^h v_j g^{''}(\wb_j.\xb) \wb_j^T \wb_j
$$

Now for an input sample $\xb$ perturbation $\deltb$, for the change in feature importance:
$$
\nabla_{\xb} S(\xb+\deltb) - \nabla_{\xb}S(\xb) \approx \nabla^2_{\xb} S \cdot \deltb 
$$
which is equal to:
$$
\sum_{j=1}^h v_j g^{''}(\wb_j.\xb) (\wb_j \cdot \deltb) \wb_j
$$
We further assume that the input is high-dimensional so that $h<d$ and for $i \neq j$ we have $\wb_j \cdot \wb_i = 0$. For maximizing the $\ell_2$ norm of feature importance difference we have the following perturbation direction:
$$
\deltb_m = \mbox{argmax}_{||\deltb||=1}||\nabla_{\xb} S(\xb+\deltb) - \nabla_{\xb}S(\xb)|| = \hat{\wb}_k
$$
where:
$$
k = \mbox{argmax}{\lvert v_j g^{''}(\wb_j.\xb)\rvert} \times ||\wb_k||^2_2
$$
comparing which to the direction of feature importance:
$$
\frac{\nabla_{\xb} S(\xb) }{||\nabla_{\xb} S(\xb)||_2}= \sum_{i=1}^h \frac{v_j g^{'}(\wb_i \cdot \xb) ||\wb_i||_2}{(\sum_{j=1}^h v_j g^{'}(\wb_j \cdot \xb) ||\wb_j||_2)^2} \hat{\wb}_i
$$
we conclude that the two directions are not parallel unless $g^{'}(.) = g^{''}(.)$ which is not the case for many activation functions like Softplus, Sigmoid, etc.

\newpage
\section{Designing interpretability-robust networks}
\label{appendix:defense}

The analyses and experiments in this paper have demonstrated that small perturbations in the input layers of deep neural networks can have large changes in the interpretations. This is analogous to classical adversarial examples, whereby small perturbations in the input produce large changes in the \textit{prediction}. In that setting, it has been proposed that the Lipschitz constant of the network be constrained during training to limit the effect of adversarial perturbations \cite{szegedy2013intriguing}. This has found some empirical success \cite{cisse2017parseval}.

Here, we propose an analogous method to upper-bound the change in interpretability of a neural network as a result of perturbations to the input. Specifically, consider a network with $K$ layers, which takes as  input a data point we denote as $y_0$. The output of the $i^\text{th}$ layer is given by $y_{i+1} = f_i(y_{i})$ for $i=0, 1 \ldots K-1$. We define $S = f_{K-1}(f_{K-2}(\ldots f_0(y_0) \ldots))$ to be the output (e.g. score for the correct class) of our network, and we are interested in designing a network whose gradient $S' = \nabla_{y_0}S$ is relatively insensitive to perturbations in the input, as this corresponds to a network whose feature importances are robust.

A natural quantity to consider is the Lipschitz constant of $S'$ with respect to $y_0$. By the chain rule, the Lipschitz constant of $S'$ is
\begin{align}
    \mathcal{L}(S') = \mathcal{L}(\frac{\delta y_k}{\delta y_{k-1}})\ldots\mathcal{L}(\frac{\delta y_1}{\delta y_{0}})
    \label{eqn:chain_rule_lipschitz}
\end{align}

Now consider the function $f_i(\cdot)$, which maps $y_{i}$ to $y_{i+1}$. In the simple case of the fully-connected network, which we consider here, $f_i(y_{i}) = g_i(W_i y_{i})$, where $g_i$ is a non-linearity and $W_i$ are the trained weights for that layer. Thus, the Lipschitz constant of the $i^{\text{th}}$ partial derivative in (\ref{eqn:chain_rule_lipschitz}) is the Lipschitz constant of 

$$\frac{\delta f_i}{\delta y_{i-1}} = W_i g_i'(W_i y_{i-1}),$$

which is upper-bounded by $\lvert \lvert W_i \rvert \rvert^2 \cdot \mathcal{L}(g_i'(\cdot))$, where $\lvert \lvert W \rvert \rvert$ denotes the operator norm of $W$ (its largest singular value)\footnote{this bound follows from the fact that the Lipschitz constant of the composition of two functions is the product of their Lipschitz constants, and the Lipschitz constant of the product of two functions is also the product of their Lipschitz constants.}. This suggests that a conservative upper ceiling for (\ref{eqn:chain_rule_lipschitz}) is

\begin{align}
    \mathcal{L}(S') \le \prod_{i=0}^{K-1} \lvert \lvert W_i \rvert \rvert^2 \mathcal{L}(g_i'(\cdot))
    \label{eqn:lipschitz_product}
\end{align}

Because the Lipschitz constant of the non-linearities $g_i'(\cdot)$ are fixed, this result suggests that a regularization based on the operator norms of the weights $W_i$ may allow us to train networks that are robust to attacks on feature importance. The calculations in this Appendix section is meant to be suggestive rather than conclusive, since in practice the Lipschitz bounds are rarely tight. 

\end{onecolumn}

\end{document}